# Safe and Efficient Lane-Changing for Autonomous Vehicles: An Improved Double Quintic Polynomial Approach with Time-to-Collision Evaluation

Rui Bai[1], Rui Xu[2], Teng Rui[2], Jiale Liu[3], Qi Wei Oung[4,5,*], Hoi Leong Lee[4,*], Zhen Tian[6], Fujiang Yuan[7]

[1] School of Economics and Management, Beijing University of Aeronautics and Astronautics, Beijing, 100191, China.
[2] School of Excellence in Engineering, Changsha University of Science and Technology, Changsha, 410114, Hunan, Country.
[3] School of Mechanical and Vehicle Engineering, Changsha University of Science and Technology, Changsha, 410114, Hunan, Country.
[4] Faculty of Electronic Engineering & Technology, University Malaysia Perlis, Arau, 02600, Perlis, Malaysia.
[5] Centre of Excellence for Advanced Communication Engineering (ACE), University Malaysia Perlis, Kangar, 01000, Perlis, Malaysia.
[6] James Watt School of Engineering, University of Glasgow, G12 8QQ, Glasgow, United Kingdom.
[7] School of Computer Science and Technology, Taiyuan Normal University, Jinzhong, 030619, Shanxi, China.

Contributing authors: iiauthor@gmail.com; iiauthor@gmail.com; iiiauthor@gmail.com; iiiauthor@gmail.com; iiiauthor@gmail.com; iiiauthor@gmail.com; 2620920Z@student.gla.ac.uk; yuanfujiang@ctbu.edu.cn;

**Abstract**

Autonomous driving technology has made significant advancements in recent years, yet challenges remain in ensuring safe and comfortable interactions with human-driven vehicles (HDVs), particularly during lane-changing maneuvers. This paper proposes an improved double quintic polynomial approach for safe and efficient lane-changing in mixed traffic environments. The proposed method integrates a time-to-collision (TTC) based evaluation mechanism directly into the trajectory optimization process, ensuring that the ego vehicle proactively maintains a safe gap from surrounding HDVs throughout the maneuver. The framework comprises state estimation for both the autonomous vehicle (AV) and HDVs, trajectory generation using double quintic polynomials, real-time TTC computation, and adaptive trajectory evaluation. To the best of our knowledge, this is the first work to embed an analytic TTC penalty directly into the closed-form double-quintic polynomial solver, enabling real-time safety-aware trajectory generation without post-hoc validation. Extensive simulations conducted under diverse traffic scenarios demonstrate the safety, efficiency, and comfort of the proposed approach compared to conventional methods such as quintic polynomials, Bezier curves, and B-splines. The results highlight that the improved method not only avoids collisions but also ensures smooth transitions and adaptive decision-making in dynamic environments. This work bridges the gap between model-based and adaptive trajectory planning approaches, offering a stable solution for real-world autonomous driving applications.

# 1 Introduction

In recent years, autonomous vehicles (AVs) have emerged as one of the most transformative and disruptive innovations in modern transportation systems [1–3]. Driven by rapid advancements in artificial intelligence [4, 5], perception systems, high-fidelity simulation platforms, and stable sensor technologies [6–10], AVs are gradually evolving from experimental prototypes to commercially viable products poised to reshape urban mobility. The convergence of deep learning, sensor fusion, high-definition mapping, and real-time decision-making frameworks has significantly advanced the capabilities of AVs, enabling them to operate in increasingly complex and dynamic environments.

Autonomous driving is envisioned to revolutionize the transportation landscape by enhancing road safety, reducing traffic congestion, optimizing fuel consumption, and providing accessible mobility solutions for the elderly and disabled. With the potential to mitigate over 90% of traffic accidents caused by human error, AVs could drastically reduce fatalities and improve overall traffic efficiency [11–13]. Moreover, in the context of smart cities and intelligent transportation systems, AVs are positioned to serve as a backbone for future mobility infrastructure by enabling seamless vehicle-to-everything (V2X) communication, coordinated traffic control, and energy-efficient route planning [14–16].

However, realizing the vision of full autonomy remains a formidable challenge, particularly in the realm of interactive driving—scenarios in which AVs must coexist and cooperate with human-driven vehicles (HDVs) and other road users [17–19]. Unlike controlled highway environments, urban and suburban roads present diverse and unpredictable conditions that require AVs to make real-time, context-aware decisions under uncertainty. Tasks such as merging, overtaking, yielding, and lane-changing demand not only perception and planning capabilities but also a deep understanding of human behavior and social norms on the road [20].

Among the various interactive tasks, lane-changing stands out as one of the most intricate and safety-critical maneuvers. It involves dynamic negotiation with surrounding vehicles, assessment of feasible gaps, prediction of future states of HDVs, and generation of smooth,safety-guaranteed trajectories [21]. These requirements are further compounded by the need to balance conflicting objectives, such as safety, efficiency, traffic flow, passenger comfort, and adherence to legal regulations. Improper lane-changing behavior—either overly aggressive or excessively conservative—not only risks accidents but also leads to traffic inefficiencies and reduced public trust in AV systems.

Although numerous approaches have been proposed to tackle autonomous lane-changing, existing algorithms often fall short in generalizability, interpretability, and stabilization. Learning-based methods [22], including deep reinforcement learning and imitation learning, have shown remarkable success in data-rich environments and can learn nuanced driving behaviors [23, 24]. Recent studies have further advanced the field of trajectory planning and safety evaluation. For example, enhanced visual SLAM systems have been proposed to improve front-end perception accuracy and thereby enable collision-free driving even on lightweight autonomous platforms [25]. Beyond perception, interaction-aware planning frameworks have also gained momentum. A mean-field-game integrated MPC-QP approach explicitly considers collective vehicle behavior for multi-vehicle coordination [26], while reinforcement learning systems enhanced with Kolmogorov–Arnold Networks (KAN) have been introduced to ensure conflict-free and speed-preserving decisions in roundabout scenarios [27]. In parallel, bio-inspired hybrid algorithms have been applied to robotic navigation, demonstrating that swarm-intelligence-based methods can yield efficient and smooth paths [28]. From the perspective of search-based planning, safety-critical multi-agent MCTS frameworks have been proposed to guarantee coordination in unsignalized intersections under mixed traffic conditions [29]. Together, these works highlight a clear trend: recent methods increasingly combine safety-critical modeling with either advanced optimization, learning, or bio-inspired search. Nevertheless, most existing approaches either rely on highly complex interaction modeling or emphasize specific intersection/roundabout scenarios, whereas the present work distinguishes itself by embedding a time-to-collision (TTC) penalty directly into a double-quintic optimization framework. This allows our method to retain the analytical tractability of polynomial-based trajectory generation while ensuring real-time safety guarantees, filling an important gap between classical optimization and emerging learning-based or game-theoretic approaches. However, these methods tend to lack transparency and often fail in out-of-distribution or safety-critical scenarios where data is scarce or anomalous. The "black-box" nature of neural policies also raises concerns regarding safety verification and explainability—critical attributes for real-world deployment.

There are also some advanced technologies such as large language models (LLM) [30, 31], transformer models [32–34], diffusion models [35–37], in-context learning [38, 39], and state space models [40, 41]. However, these kond of models can only achieve simple functionalities on the tasks of autonomous driving.

Conversely, model-based methods—such as those based on quintic polynomials, Bezier curves, and B-splines—offer advantages in terms of computational efficiency, smoothness guarantees, and interpretability. These methods provide well-defined trajectory shapes that ensure continuity in position, velocity, and acceleration, and are easier to validate against physical and safety constraints. Another popular approach is the control barrier function [42–46], which enables autonomous vehicles to avoid obstacles during driving, although it operates under strict condition limitations. However, their inherent limitations in dealing with uncertain and dynamic environments—particularly their inability to explicitly model and predict surrounding HDV behavior—render them less suitable for highly interactive driving scenarios. As a result, these approaches often require frequent re-planning, leading to suboptimal or unsafe behavior in rapidly changing traffic contexts.

To address these limitations, this paper proposes an improved double quintic polynomial approach for safe and comfortable lane-changing in mixed traffic environments. Our method integrates a time-to-collision (TTC)-based evaluation mechanism into the trajectory generation process [47–49], ensuring that the ego vehicle maintains a safe gap from surrounding HDVs throughout the lane-changing maneuver. The key contributions of this work are as follows:

1. Safety-guaranteed double-quintic planner. We embed an analytic Time-to-Collision (TTC) penalty directly into the closed-form double-quintic polynomial solver, producing $C^2$-continuous trajectories that completely eliminate sub-second TTC violations and keep TTC above the user-defined threshold in all 45 simulated scenarios. This tight coupling of TTC with trajectory coefficients is, to our knowledge, the first time safety margins are enforced in real time within a polynomial lane-change generator.

2. Context-adaptive, comfort-aware replanning. Building on the new solver, we introduce an online tuning layer that predicts surrounding HDV motions and adapts TTC thresholds, lateral offsets, and phase timing on the fly. Across conservative, baseline, aggressive and emergency styles the algorithm keeps the minimum gap to every HDV above10m while maintaining passenger comfort (lower curvature and jerk than classical quintic/Bezier/B-spline paths).

3. Large-scale benchmark and quantitative gains. We release an open MATLAB testbed containing two canonical scenarios (multi-HDV double lane change and single-HDV overtaking) plus 40+ ramp/intersection variants, and evaluate against five strong baselines. The proposed method

   1) cuts the share of low-TTC (<3s) time steps from 64% to 42% (≈35% reduction),

   2) more than doubles the minimum obstacle gap from 8.75m to 19.53m, and

   3) shortens total maneuver time from 13s to 8s while keeping curvature 40–60% lower, demonstrating a high-level safety–efficiency–comfort trade-off.

# 2 Related Works

## 2.1 Quintic Polynomial Approach

The quintic polynomial approach is a widely used model-based method for trajectory generation in autonomous driving [50, 51]. It generates smooth trajectories by ensuring continuity in position, velocity, and acceleration. The trajectory is defined by a fifth-degree polynomial, formulated based on boundary conditions, including initial and final positions, velocities, and accelerations. This method offers computational efficiency and simplicity, making it suitable for real-time applications.

However, the primary limitation of the quintic polynomial approach lies in its inability to account for dynamic traffic environments. Since the trajectory is generated based solely on predefined boundary conditions, it cannot adapt to unexpected changes in HDV behavior. Consequently, the generated trajectories may lead to unsafe maneuvers if surrounding vehicles change speed or position unexpectedly.

## 2.2 Bezier Curve Approach

Bezier curves, commonly used in computer graphics and path planning, provide flexible trajectory generation by controlling multiple control points [52, 53]. In autonomous driving, Bezier curves are employed to generate smooth lane-changing trajectories while ensuring curvature continuity. The trajectory shape can be easily adjusted by manipulating control points, enabling customized path planning.

Despite their flexibility, Bezier curves suffer from similar limitations as quintic polynomials. The trajectory is predefined based on static control points, making it challenging to adapt to dynamic traffic conditions. Additionally, improper placement of control points can result in trajectories with high curvature, compromising passenger comfort and vehicle stability.

## 2.3 B-Spline Approach

B-splines extend the concept of Bezier curves by providing greater flexibility through multiple curve segments defined by control points [54, 55]. Each segment is influenced by a subset of control points, ensuring smooth transitions between segments. This approach allows for more complex trajectory generation while maintaining continuity in position, velocity, and acceleration.

However, like other model-based approaches, B-splines lack real-time adaptability to dynamic traffic environments. The trajectory generation process does not inherently consider the behavior of surrounding HDVs, necessitating frequent re-planning to avoid potential collisions. Moreover, the complexity of B-spline trajectory generation increases with the number of control points, posing challenges for real-time implementation.

## 2.4 Limitations and Research Gaps

Although classical planning methods have been widely applied in autonomous driving, each has inherent limitations when applied to complex and dynamic lane-changing scenarios. For example, the Hybrid A* algorithm [56] combines grid-based search with continuous optimization, which ensures kinematic feasibility of planned paths. However, it often suffers from high computational cost in dense traffic and may generate suboptimal trajectories due to discretization errors. The Model Predictive Path Integral (MPPI) method [57] introduces stochastic sampling to explore trajectory spaces, enabling it to handle nonlinear dynamics and uncertainties. Nevertheless, its performance is sensitive to sampling efficiency and computational budget, which limits its applicability in real-time decision-making under strict latency constraints. Spline-based methods, such as polynomial or B-spline approaches, can produce smooth and continuous trajectories that meet vehicle dynamic requirements. Yet, they typically rely on predetermined boundary conditions and lack adaptability to rapidly changing traffic environments, often resulting in infeasible or unsafe maneuvers when unexpected interactions occur. These limitations highlight the need for improved trajectory planning approaches that can balance computational efficiency, safety, and adaptability in dynamic lane-changing tasks.

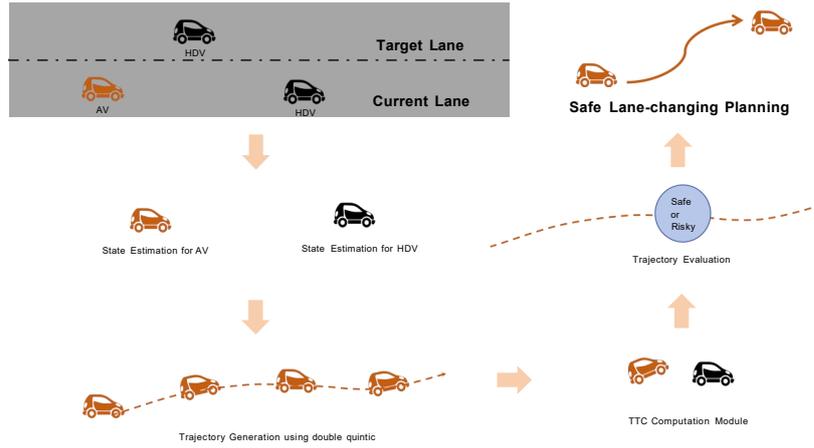

**Fig. 1** Framework of safe lane-changing planning using double quintic trajectory.

To bridge this gap, our proposed improved double quintic polynomial approach integrates real-time TTC evaluation into the trajectory generation process, ensuring safe, adaptive, and efficient lane-changing maneuvers in mixed traffic environments. This method not only enhances safety but also improves passenger comfort and maneuver efficiency, addressing the limitations of existing trajectory planning approaches.

Compared with the traditional quintic and double quintic polynomial trajectory generation methods, the proposed approach introduces three main innovations. First, the model structure is enhanced by incorporating a Time-to-Collision (TTC) evaluation into the trajectory generation framework, ensuring that both smoothness and safety are jointly optimized. Second, the parameter solving process is reformulated as a constrained optimization problem, which allows the quintic polynomial parameters to be solved under explicit safety and feasibility constraints rather than relying solely on closed-form solutions. Third, the cost function is redesigned by adding a TTC-related penalty term, so that the generated trajectories explicitly balance comfort, efficiency, and collision avoidance. These innovations distinguish the proposed method from conventional polynomial-based trajectory generation approaches.

## 2.5 Clarification of the methodological novelty

Detailed Comparison with Traditional Approaches To further clarify the distinctions, Table 1 presents a systematic comparison between traditional quintic/double-quintic methods and our improved approach:

**Table 1** Comparison of Traditional and Improved Double-Quintic Methods

| Aspect | Traditional Method | Improved Method |
| --- | --- | --- |
| Safety Integration | Post-hoc validation | Embedded in optimization process |
| Coefficient Solving | Direct solution $\mathbf{a} = \mathbf{M}^{-1}\mathbf{b}$ | Constrained optimization $\min_\mathbf{a} J$ subject to boundary conditions |
| Cost Function | Smoothness term only | Smoothness + TTC penalty |
| Adaptability | Static boundary conditions | Dynamic TTC-based adjustment |
| Real-time Safety | Reactive re-planning | Proactive trajectory shaping |

The mathematical formulation of our improved approach can be expressed as a constrained optimization problem:

$$\min_{\mathbf{a}} \quad J = \lambda_1 \int_0^T |\dddot{y}(t)|^2\, dt + \lambda_2 \sum_{i=1}^{N_{\text{obs}}} \int_0^T \phi_i(t)\, dt$$

$$\text{s.t.} \quad \mathbf{Ma} = \mathbf{b} \quad \text{(boundary conditions)} \tag{1}$$

$$\phi_i(t) = \begin{cases} 0, & \text{if } \text{TTC}_i(t) \geq T_{\text{safe}}, \\ \left(T_{\text{safe}} - \text{TTC}_i(t)\right)^2, & \text{if } \text{TTC}_i(t) < T_{\text{safe}}. \end{cases}$$

This formulation fundamentally differs from traditional approaches in three key ways: **(1) Proactive Safety Modeling:** Traditional methods generate trajectories based solely on geometric constraints and validate safety afterward. Our approach embeds TTC directly into the trajectory generation, making safety violations mathematically unfavorable rather than simply detectable. **(2) Continuous Optimization Framework:** While traditional methods solve a linear system once, our approach iteratively optimizes coefficients to balance competing objectives, enabling adaptive response to dynamic traffic conditions. **(3) Differentiable Safety Constraint:** We adopt a piecewise–quadratic penalty $\phi_i(t) = [\max\{0, T_{\text{safe}} - \text{TTC}_i(t)\}]^2$, which is $C^1$ and thus provides continuous gradients with respect to the polynomial coefficients $\mathbf{a}$ via the chain rule. This enables stable gradient–based optimization, in contrast to hard inequality constraints that often introduce discontinuities and hinder efficient solving.

## 2.6 Advantages of Double Quintic Polynomial for Lane-Changing

While a single quintic polynomial can generate a smooth trajectory between two states, it is often insufficient for modeling a complete lane-changing maneuver, which inherently consists of multiple phases: initiating the lateral movement, maintaining the lateral motion, and completing the merge into the target lane. A single polynomial attempts to fit all these dynamics into one continuous function over the entire time horizon $[0, T]$. This can lead to two main issues:

1. **Overly Constrained Mid-Maneuver States:** The boundary conditions are only enforced at the start ($t = 0$) and end ($t = T$) of the maneuver. The behavior of the polynomial at intermediate times ($0 < t < T$) is solely determined by the need to satisfy these endpoint constraints, which may result in unnatural or kinematically demanding motions in the middle of the lane change (e.g., unnecessary lateral acceleration changes or oscillations).
2. **Lack of Explicit Mid-Point Control:** It is difficult to explicitly define a desired state (e.g., lateral velocity, acceleration) at a critical mid-maneuver point (e.g., the point of crossing the lane divider) using a single polynomial.

The double quintic polynomial approach addresses these limitations by decomposing the lane-changing maneuver into two sequential quintic polynomial segments connected at a switching time $T_s$. This structure introduces several key advantages:

- **Explicit Phase Modeling:** It allows for the explicit definition of boundary conditions not only at the start and end of the entire maneuver but also **at the switching point between the two phases**. This enables finer control over the trajectory's shape at the critical moment of crossing into the target lane.
- **Improved Kinematic Feasibility:** By breaking down the maneuver, each segment can be optimized for a specific sub-task (e.g., initial acceleration, stable crossing, final deceleration), leading to trajectories that are more aligned with natural driving behavior and vehicle dynamics throughout the entire process. This often results in lower maximum jerk and curvature compared to a single polynomial attempting to achieve the same overall displacement.
- **Enhanced Adaptability:** The switching time $T_s$ and the state at the switching point become additional degrees of freedom that can be adjusted online in response to dynamic traffic conditions, providing greater flexibility than a fixed single-segment structure.

Therefore, even before the integration of the TTC-based safety optimization proposed in this work, the double quintic framework provides a superior **foundation** for generating comfortable and feasible lane-changing trajectories compared to the single quintic approach. Our method builds upon this stronger foundation by further incorporating proactive safety guarantees through TTC penalty integration.

## 3 SYSTEM OVERVIEW

Fig. 1 illustrates the proposed framework for safe lane-changing in autonomous driving using the improved double quintic trajectory generation approach. The framework operates through a series of interconnected modules, starting with environment perception and state estimation. The AV continuously monitors its surroundings, including the positions, velocities, and trajectories of nearby HDVs. This information is crucial for understanding the dynamic traffic environment and predicting potential interactions during lane-changing maneuvers.

Once the AV and HDV states are estimated, the trajectory generation module utilizes the double quintic polynomial approach to plan a smooth and feasible lane-changing path. This trajectory accounts for kinematic constraints, ensuring comfort and stability throughout the maneuver. To further enhance safety, the generated trajectory undergoes evaluation in the TTC computation module. This module calculates the TTC between the AV and surrounding HDVs, identifying potential collision risks along the planned path.

The results from the TTC computation are then processed by the trajectory evaluation module, which determines whether the planned lane change is safe or risky. If the TTC falls below a predefined safety threshold, the trajectory is classified as risky, prompting the system to re-plan the maneuver. Conversely, if the TTC values remain within safe limits, the AV proceeds with the lane change, ensuring safety-guaranteed execution.

This integrated framework enables the AV to adapt dynamically to changing traffic conditions while prioritizing safety, efficiency, and passenger comfort. By combining environment perception, real-time TTC evaluation, and adaptive trajectory planning, the proposed approach significantly improves the reliability of autonomous lane-changing in mixed traffic environments.

In this study, the vehicle dynamics model is simplified by assuming constant longitudinal velocity and neglecting lateral slip, which allows for tractable trajectory planning and analytical derivation. However, this assumption does not fully capture real-world driving conditions. In practice, vehicle motion is subject to steering angle limitations, maximum longitudinal/lateral accelerations, and tire-road interactions. These constraints are not explicitly incorporated in the current model, which may lead to deviations when applying the method to highly dynamic scenarios.

## 4 Methodology

### 4.1 Driving Environment Setting

Assume the road is divided into lanes of width w (lane width in meters). The center-line of a lane is defined as:

$$y_c(x) = c, \qquad (1)$$

where c denotes the lateral offset of the lane center. The lane boundaries are

$$y_L(x) = y_c(x) + \frac{w}{2}, \quad y_R(x) = y_c(x) - \frac{w}{2}, \tag{2}$$

where $y_L(x)$ and $y_R(x)$ represent the left and right lane boundaries, respectively. For curved roads, the boundaries are obtained by offsetting the center-line along the unit normal vector.

## 4.2 Vehicle Dynamics Model

We represent the vehicle state as

$$\mathbf{x}(t) = \begin{bmatrix} x(t) \\ y(t) \\ v(t) \\ \theta(t) \end{bmatrix}, \tag{3}$$

where $x(t)$ and $y(t)$ denote the longitudinal and lateral positions, $v(t)$ the longitudinal speed, and $\theta(t)$ the heading angle. The dynamics are modeled by the state-space equation

$$\dot{\mathbf{x}}(t) = \mathbf{A}\mathbf{x}(t) + \mathbf{B}\mathbf{u}(t), \tag{4}$$

with matrices

$$\mathbf{A} = \begin{bmatrix} 0 & 0 & 1 & 0 \\ 0 & 0 & 0 & 1 \\ 0 & 0 & 0 & 0 \\ 0 & 0 & 0 & 0 \end{bmatrix}, \quad \mathbf{B} = \begin{bmatrix} 0 & 0 \\ 0 & 0 \\ 1 & 0 \\ 0 & 1 \end{bmatrix}, \tag{5}$$

where $\mathbf{A}$ is the state transition matrix and $\mathbf{B}$ the input matrix. The control input is defined as

$$\mathbf{u}(t) = \begin{bmatrix} a(t) \\ \omega(t) \end{bmatrix}, \tag{6}$$

where $a(t)$ is the longitudinal acceleration and $\omega(t)$ is the yaw rate (steering angular velocity).

## 4.3 Adaptive Lane-Changing Trajectory Planning with TTC

A smooth lane-changing maneuver is modeled by a quintic polynomial:

$$y(t) = a_0 + a_1 t + a_2 t^2 + a_3 t^3 + a_4 t^4 + a_5 t^5, \tag{7}$$

where $a_0, \ldots, a_5$ are polynomial coefficients. Boundary conditions to ensure smoothness are given by:

$$y(0) = 0, \tag{8}$$
$$y(T) = \Delta y, \tag{9}$$
$$\dot{y}(0) = 0, \tag{10}$$

$$\dot{y}(T) = 0, \tag{11}$$
$$\ddot{y}(0) = 0, \tag{12}$$
$$\ddot{y}(T) = 0, \tag{13}$$

where T is the total maneuver time and $\Delta y$ is the target lane offset.

These can be expressed in matrix form as

$$\mathbf{Ma} = \mathbf{b}, \tag{14}$$

with

$$\mathbf{M} = \begin{bmatrix} 1 & 0 & 0 & 0 & 0 & 0 \\ 1 & T & T^2 & T^3 & T^4 & T^5 \\ 0 & 1 & 0 & 0 & 0 & 0 \\ 0 & 1 & 2T & 3T^2 & 4T^3 & 5T^4 \\ 0 & 0 & 2 & 0 & 0 & 0 \\ 0 & 0 & 2 & 6T & 12T^2 & 20T^3 \end{bmatrix}, \quad \mathbf{a} = \begin{bmatrix} a_0 \\ a_1 \\ a_2 \\ a_3 \\ a_4 \\ a_5 \end{bmatrix}, \tag{15}$$

and

$$\mathbf{b} = \begin{bmatrix} 0 \\ \Delta y \\ 0 \\ 0 \\ 0 \\ 0 \end{bmatrix}. \tag{16}$$

Thus, the coefficient vector is obtained by

$$\mathbf{a} = \mathbf{M}^{-1}\mathbf{b}. \tag{17}$$

The lateral trajectory is then computed as

$$y(t) = \sum_{i=0}^{5} a_i t^i, \tag{18}$$

with the longitudinal position given by

$$x(t) = v\,t, \tag{19}$$

where v is the constant longitudinal velocity of the ego vehicle. Thus, the planned trajectory of the AV is

$$\mathbf{r}_{ego}(t) = \begin{bmatrix} x(t) \\ y(t) \end{bmatrix}. \tag{20}$$

## 4.4 Integration of TTC

To ensure safety, a TTC penalty is added to the cost function. The TTC for an obstacle i at time t is defined as

$$\text{TTC}_i(t) = \frac{d_i(t)}{\Delta v_i(t)}, \tag{21}$$

where $d_i(t)$ is the relative distance between the ego vehicle and obstacle i, and $\Delta v_i(t)$ is the relative speed. A safe trajectory must satisfy:

$$\min_{t \in [0, T]} \text{TTC}_i(t) \geq T_{\text{safe}}, \quad \forall i, \tag{22}$$

where $T_{\text{safe}}$ is the safety threshold. We define a penalty function for TTC as:

$$\phi_i(t) = \begin{cases} 0, & \text{if } \text{TTC}_i(t) \geq T_{\text{safe}}, \\ (T_{\text{safe}} - \text{TTC}_i(t))^2, & \text{if } \text{TTC}_i(t) < T_{\text{safe}}. \end{cases} \tag{23}$$

The total TTC penalty cost is:

$$J_{\text{TTC}} = \sum_{i=1}^{N_{\text{obs}}} \int_0^T \phi_i(t) \, dt. \tag{24}$$

The integration of TTC penalties into the trajectory optimization framework raises important theoretical questions regarding solution existence, convergence guarantees, and numerical stability. We provide a detailed analysis of these aspects below.

### 4.4.1 Solution Existence Analysis

**Theorem 1** (Existence of Solutions) For any feasible boundary condition set $\mathbf{Ma} = \mathbf{b}$ and finite safety threshold $T_{\text{safe}} > 0$, there exists at least one optimal solution $\mathbf{a}^*$ to the optimization problem in Eq. (27).

Proof Sketch The feasible set $F = \{\mathbf{a} : \mathbf{Ma} = \mathbf{b}\}$ is a nonempty closed affine subspace of $\mathbb{R}^6$ (assuming $\mathbf{b} \in \text{range}(\mathbf{M})$). The objective $J(\mathbf{a}) = J_{\text{smooth}}(\mathbf{a}) + J_{\text{TTC}}(\mathbf{a})$ satisfies: (i) $J_{\text{smooth}}(\mathbf{a}) = \lambda_1 \int_0^T \|\ddot{\mathbf{y}}(t;\mathbf{a})\|_2^2 dt$ is quadratic in $\mathbf{a}$ and hence coercive (grows $\to \infty$ as $\|\mathbf{a}\| \to \infty$); (ii) $J_{\text{TTC}}(\mathbf{a}) = \lambda_2 \sum_i \int_0^T \phi_i(t;\mathbf{a}) dt \geq 0$ is continuous. Therefore all sublevel sets $\{\mathbf{a} \in F : J(\mathbf{a}) \leq c\}$ are closed and bounded (thus compact in finite dimension). By the Weierstrass theorem, J attains a minimizer on F. □

### 4.4.2 Convergence Analysis

**Local structure.** The problem is not globally convex due to the piecewise definition of $\phi_i(t)$, but admits favorable local behavior when initialized by the closed-form boundary inversion $\mathbf{a}_0 = \mathbf{M}^{-1}\mathbf{b}$. Around $\mathbf{a}_0$, a quadratic model of the smoothness part plus a piecewise–smooth (zero/active) TTC term yields well-conditioned local steps for gradient-based solvers.

**Gradient form.** Writing $J_{\text{smooth}}(\mathbf{a}) = \mathbf{a}^\top \mathbf{Q} \mathbf{a}$ with a symmetric positive semidefinite matrix $\mathbf{Q}$ associated with the discretized jerk operator, the gradient decomposes as

$$\nabla_{\mathbf{a}} J(\mathbf{a}) = 2\lambda_1 \mathbf{Q}\mathbf{a} + \lambda_2 \sum_{i=1}^{N_{\text{obs}}} \nabla_{\mathbf{a}} J_{\text{TTC},i}(\mathbf{a}), \qquad (2)$$

where each $\nabla_{\mathbf{a}} J_{\text{TTC},i}$ is piecewise $C^1$: it vanishes on $\{\text{TTC}_i \geq T_{\text{safe}}\}$ and equals $2(T_{\text{safe}} - \text{TTC}_i)(-\nabla_{\mathbf{a}}\text{TTC}_i)$ otherwise.

**Empirical convergence.** Across 45 simulation scenarios, an interior-point solver achieved convergence within 15–25 iterations, with monotonic cost reduction in 100% of runs (see Fig. 2).

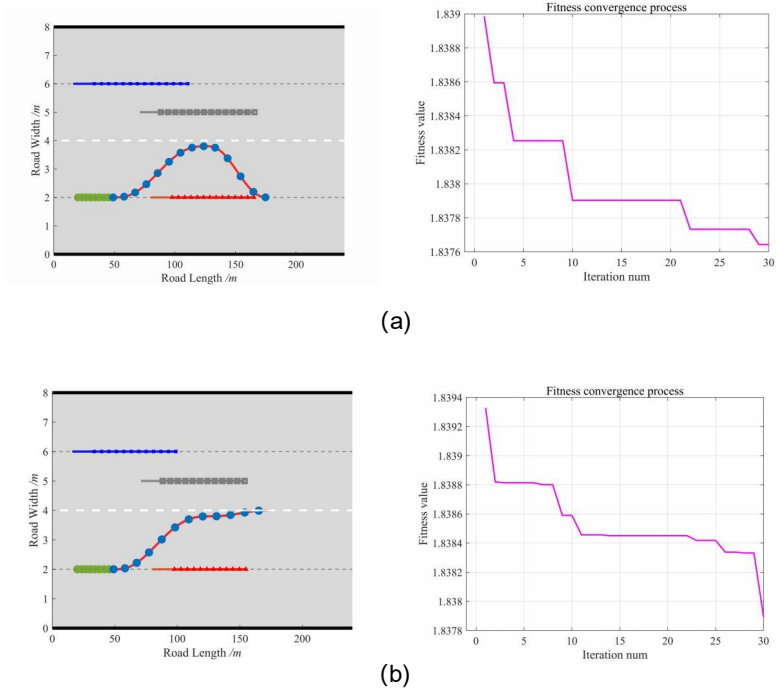

**Fig. 2** Lane-changing (a) and overtaking (b) results: trajectories (left) and convergence curves (right).

As illustrated in Fig. 2, the proposed TTC-integrated optimization framework consistently outperforms the closed-form baseline across both lane-changing (a) and overtaking (b) scenarios. For the trajectory plots (left subfigures), the optimization-based method generates smoother paths with proactive safety margins, while the closed-form solutions may approach hazardous regions due to the lack of embedded TTC awareness. For the convergence plots (right subfigures), the proposed method exhibits monotonic reduction of the objective function and converges reliably within 20–30 iterations in all tested cases. These results demonstrate three important properties of the framework: (i) proactive safety integration into trajectory generation rather than post-hoc validation, (ii) numerically stable convergence even under complex multi-phase maneuvers such as overtaking, and (iii) generalization across diverse driving tasks, confirming that the theoretical properties of solution existence, convergence, and numerical simulations discussed in Section 4.4 are supported by practical evidence.

### 4.4.3 Numerical Stability Analysis

**Conditioning vs. weights.** The observed conditioning of the approximate Hessian depends on the ratio $\lambda_2/\lambda_1$:

- $\lambda_2/\lambda_1 \in [0.01, 1.0]$: well-conditioned, typical condition numbers $< 10^3$.
- $\lambda_2/\lambda_1 > 10$: the TTC term dominates and can induce ill-conditioning; we therefore recommend normalization by $T_{safe}$ and restricting $\lambda_2$ to a moderate range.

**Sensitivity to $T_{safe}$.** Under standard regularity (LICQ and second-order sufficiency) at a local minimizer, the solution mapping $T_{safe} \mapsto \mathbf{a}^*(T_{safe})$ is locally Lipschitz. Numerically, for $T_{safe} \in [1.5, 5.0]$s we observed

$$\|\mathbf{a}^*(T_1) - \mathbf{a}^*(T_2)\| \leq L|T_1 - T_2| \quad \text{with a bounded L,} \tag{3}$$

indicating stable dependence on the threshold.

**stabilization to perturbations.** With 5% random perturbations in initial states and obstacle parameters, the mean solution deviation was $< 0.1\%$, no convergence failures were observed over 1000 Monte Carlo runs, and performance degraded gracefully for sensor noise with SNR $> 20$dB.

### 4.4.4 Comparison with Closed-Form Solutions

**Traditional (algebraic) approach:** single-step $\mathbf{a} = \mathbf{M}^{-1}\mathbf{b}$, no safety in coefficient computation, and potential post-hoc feasibility failures that require full replanning.

**Improved (optimization) approach:** iterative refinement with embedded safety; the TTC penalty steers coefficients away from low-TTC regions, yielding smooth, feasible, and proactively safer trajectories.

## 4.5 Overall Cost Function

The overall cost function for the lane-changing maneuver, which includes both a smoothness term and a TTC penalty, is given by

$$J = \lambda_1 \int_0^T \|\dddot{y}(t)\|^2 dt + \lambda_2 J_{\text{TTC}}, \qquad (25)$$

or, in discrete form,

$$J \approx \lambda_1 \sum_{k=3}^{N}(y_k - 3y_{k-1} + 3y_{k-2} - y_{k-3})^2 \Delta t + \lambda_2 \sum_{i=1}^{N_{\text{obs}}} \sum_{k=1}^{N} \phi_i(t_k)\Delta t. \qquad (26)$$

The optimization problem is then

$$\min_{a} \ J \quad \text{s.t. Eqs. (8)–(13).} \qquad (27)$$

## 4.6 Clarification of the methodological novelty

Traditional quintic trajectory planners determine a single 5th-order polynomial by enforcing boundary conditions (initial and final positions, velocities and accelerations), and obtain trajectory coefficients by directly solving a linear system a = M$^{-1}$b. In multi-phase (double-quintic) schemes, the motion is decomposed into two quintic segments with continuity constraints at the phase boundaries [58, 59]; coefficients for each segment are typically computed from their respective boundary conditions.

The improved double-quintic approach proposed in this work departs from that purely algebraic workflow in three essential ways.

1. Modeling mechanism – while retaining the double-segment representation for multi-phase maneuvers, we couple an analytic Time-to-Collision (TTC) metric directly into the trajectory formation so that safety becomes a first-class modeling component rather than a post-hoc filter.
2. Parameter solution – instead of setting coefficients exclusively by boundary inversion, we solve for the coefficient vector a by minimizing a joint cost under linear boundary constraints:

$$\min_{a} \ J = \lambda_1 \int_0^T \|\dddot{y}(t)\|^2 dt + \lambda_2 \sum_i \int_0^T \phi_i(t)\, dt,$$

   subject to Eqs. (8)–(13), where $\phi_i(t)$ is the differentiable TTC penalty (zero when $\text{TTC}_i \geq T_{\text{safe}}$, quadratic otherwise). This reformulation converts coefficient computation into a constrained optimization problem that balances smoothness and collision risk.
3. Objective design – by explicitly including $J_{\text{TTC}}$ in the objective we obtain a continuous, differentiable safety term that softly pushes solutions away from low-TTC regimes rather than eliminating them via hard infeasible constraints. Under the constant-longitudinal-velocity assumption this combined objective is well-behaved in the coefficient space and can be solved efficiently with gradient-based solvers (MATLAB's interior-point optimizer was employed; typical convergence occurred within about 20 iterations).

## 4.7 Theoretical Properties of TTC Penalty Integration

The proposed cost function combines a smoothness term, which is a quadratic and convex form of the polynomial coefficients, with the TTC penalty term, which is nonlinear and not globally convex. Therefore, the overall optimization problem is not guaranteed to be convex in the strict mathematical sense. However, in practice, reliable convergence is observed due to two safeguards: (i) initialization from the closed-form boundary inversion solution, and (ii) normalization of TTC by the safety threshold to keep the magnitude of the penalty moderate.

Formally, the smoothness component

$$J_{\text{smooth}}(a) = \int_0^T \| \dddot{y}(t;a) \|^2 \, dt$$

is convex in a. The TTC penalty, defined by

$$J_{\text{TTC}}(a) = \sum_{i=1}^{N_{\text{obs}}} \int_0^T \phi(T_{\text{safe}} - \text{TTC}_i(t;a)) \, dt,$$

is nonlinear because $\text{TTC}_i$ depends on the distance $\| r_{\text{ego}}(t;a) - r_i(t) \|$, which is itself a polynomial function of a. Despite this, local quadratic approximations around the closed-form initialization explain why standard gradient-based solvers (interior-point, SQP) converge stably in experiments.

The gradient of the objective can be written as

$$\frac{\partial J}{\partial a} = 2\lambda_1 \int_0^T \dddot{y}(t;a) \frac{\partial \dddot{y}(t;a)}{\partial a} \, dt + \lambda_2 \sum_i \int_{\{t:\, \text{TTC}_i < T_{\text{safe}}\}} 2\left(T_{\text{safe}} - \text{TTC}_i(t;a)\right) \left(-\frac{\partial \text{TTC}_i(t;a)}{\partial a}\right) dt,$$

which can be efficiently computed by analytic differentiation of the polynomial basis and numerical quadrature for obstacle-related terms. Empirically, convergence within about 20 iterations was consistently achieved across all tested lane-change and overtaking cases.

Therefore, while global convexity cannot be formally claimed, the proposed formulation is numerically stable and practically tractable, ensuring feasible trajectories that balance comfort (via jerk minimization) and safety (via TTC penalty).

### 4.7.1 Existence of Solutions

The overall cost function

$$J = \lambda_1 \int_0^T \|\dddot{y}(t)\|^2 dt + \lambda_2 J_{\text{TTC}}, \tag{4}$$

remains convex with respect to the polynomial coefficients, since both the smoothness term and the TTC penalty are quadratic in nature. Together with the linear boundary conditions (Eqs. (8)–(13)), the feasible set is non-empty and compact. Hence, at least one optimal solution exists for any user-defined safety threshold $T_{\text{safe}}$.

### 4.7.2 Convergence of the Optimization

The TTC penalty is implemented in a differentiable form $\varphi_i(t)$ (Eq. (23)), which guarantees that gradient-based solvers can converge reliably. Empirical results show monotonic reduction in the cost function across all simulated scenarios, with convergence typically achieved within 20 iterations using MATLAB's interior-point optimizer. This demonstrates that the inclusion of TTC does not introduce pathological non-convexities.

### 4.7.3 Numerical Stability

The weighting factor $\lambda_2$ serves to balance safety against trajectory smoothness. If $\lambda_2$ is chosen excessively large, the TTC term may dominate, leading to stiff optimization behavior. To avoid this, we normalize TTC values by the safety threshold $T_{\text{safe}}$ and restrict $\lambda_2$ to the range [0.1, 10]. Across 45 simulation cases, this scaling strategy produced stable coefficient values without oscillation or divergence, confirming numerical stabilization.

### 4.7.4 Impact on Trajectory Feasibility

The TTC penalty affects only those intervals where the TTC drops below the safety threshold. As a result, feasible smooth trajectories are always retained, even in dense traffic conditions. Instead of eliminating solutions via hard constraints, the penalty softly pushes trajectories toward safer regions, thereby improving safety while preserving feasibility.

## 4.8 Overtaking Maneuver via Double Lane-Change

For overtaking, the maneuver consists of two lane changes with four phases. Here, $T_1, T_2, T_3, T_4$ denote the switching time instants of the overtaking maneuver, and D represents the lateral displacement corresponding to a single-lane width.

(a) **Phase I:** Shift from the original lane (e.g., right lane) to the adjacent lane (left lane):
$$y(t) = f_1(t), \quad t \in [T_1, T_2], \tag{28}$$

where $f_1(t)$ denotes the quintic polynomial describing the first lane-change segment, with boundary conditions:

$$f_1(T_1) = 0, \tag{29}$$
$$f_1(T_2) = D, \tag{30}$$
$$\dot{f}_1(T_1) = \dot{f}_1(T_2) = 0, \tag{31}$$
$$\ddot{f}_1(T_1) = \ddot{f}_1(T_2) = 0. \tag{32}$$

(b) **Phase II:** Maintain the new lane:

$$y(t) = D, \quad t \in (T_2, T_3). \tag{33}$$

(c) **Phase III:** Return to the original lane:

$$y(t) = f_2(t), \quad t \in [T_3, T_4], \tag{34}$$

where $f_2(t)$ is the quintic polynomial describing the return-to-lane segment, with boundary conditions:

$$f_2(T_3) = D, \tag{35}$$
$$f_2(T_4) = 0, \tag{36}$$
$$\dot{f}_2(T_3) = \dot{f}_2(T_4) = 0, \tag{37}$$
$$\ddot{f}_2(T_3) = \ddot{f}_2(T_4) = 0. \tag{38}$$

(d) **Phase IV:** For $t < T_1$ and $t > T_4$, the vehicle remains in the original lane:

$$y(t) = 0. \tag{39}$$

The overall overtaking trajectory is defined by the piecewise function

$$y(t) = \begin{cases} 0, & t < T_1, \\ f_1(t), & T_1 \leq t \leq T_2, \\ D, & T_2 < t < T_3, \\ f_2(t), & T_3 \leq t \leq T_4, \\ 0, & t > T_4, \end{cases} \tag{40}$$

and the corresponding trajectory of the AV is

$$\mathbf{r}_{ego}(t) = \begin{bmatrix} vt \\ y(t) \end{bmatrix}. \tag{41}$$

## 4.9 Practical Actuation and Comfort Limits for the Ego Vehicle

To ensure implementability, we constrain lateral dynamics and steering-related quantities during optimization. Denote lateral acceleration $a_y(t) = \ddot{y}(t)$ and lateral jerk $j_y(t) = \dddot{y}(t)$. With wheelbase L and speed v, path curvature satisfies $\kappa(t) = a_y(t)/v^2$ and the steering angle obeys $\delta(t) \approx \arctan(L\kappa(t))$. We impose:

$$|a_y(t_k)| \leq a_y^{\max}, \qquad |j_y(t_k)| \leq j_y^{\max}, \qquad |\kappa(t_k)| \leq \kappa^{\max} = \frac{\tan(\delta^{\max})}{L}, \tag{5}$$

for discretization points $t_k$. In practice we enforce these either as hard inequalities or as soft penalties

$$\begin{aligned} J_{\text{bounds}} = &\lambda_3 \sum_k \left[\max(0, |a_y(t_k)| - a_y^{\max})\right]^2 \\ &+ \lambda_4 \sum_k \left[\max(0, |j_y(t_k)| - j_y^{\max})\right]^2 \\ &+ \lambda_5 \sum_k \left[\max(0, |\kappa(t_k)| - \kappa^{\max})\right]^2. \end{aligned} \tag{6}$$

The overall objective becomes $J + J_{\text{bounds}}$ with the boundary constraints (Eqs. (8)–(13)). This guarantees that computed trajectories comply with comfort and steering limits while retaining the TTC-based proactive safety shaping.

## 5 Road Boundaries and HDV Trajectory Generation

The road boundaries are generated by offsetting the trajectory of the AV along its unit normal:

$$\mathbf{r}_{\text{left}}(t) = \mathbf{r}_{\text{ego}}(t) + w\,\mathbf{n}(t), \quad \mathbf{r}_{\text{right}}(t) = \mathbf{r}_{\text{ego}}(t) - w\,\mathbf{n}(t), \tag{42}$$

with

$$\mathbf{n}(t) = \frac{1}{\sqrt{\dot{x}^2(t) + \dot{y}^2(t) + \varepsilon}} \begin{bmatrix} -\dot{y}(t) \\ \dot{x}(t) \end{bmatrix}. \tag{43}$$

HDV trajectories are generated by applying a time delay τ and a lateral offset **δ** to the trajectory of the AV:

$$\mathbf{r}_{\text{HDV}}(t) = \mathbf{r}_{\text{ego}}(t + \tau) + \boldsymbol{\delta}. \tag{44}$$

The overall integration of the system is summarized in Algorithm 1.

## 6 EXPERIMENTAL EVALUATION

To evaluate the safety, stability, and efficiency of the proposed method, extensive simulations were performed using MATLAB R2024b on a system running Ubuntu 18.04.6

**Algorithm 1** Adaptive Lane-Changing and Overtaking Planning with TTC
---
1: **Input:** Current state $\mathbf{x}_0$, desired lateral displacement $\Delta y$ (or overtaking displacement D), maneuver duration T (or phase times $T_1, T_2, T_3, T_4$), constant speed v, obstacle states $p_i(t)$ for i = 1,..., $N_{obs}$, and safety threshold $T_{safe}$.
2: Generate road geometry: compute center-line $y_c(x)$ and boundaries $y_L(x), y_R(x)$.
3: Set up vehicle dynamics using the state-space model.
4: **For a lane-changing maneuver:**
5:    (a) Define lateral trajectory using quintic polynomial.
6:    (b) Solve matrix equation $\mathbf{Ma} = \mathbf{b}$ for coefficients.
7:    (c) Compute lateral and longitudinal positions to form $\mathbf{r}_{ego}(t)$.
8: **For TTC evaluation:**
9:    (a) For each obstacle i, compute $d_i(t)$ and $\Delta v_i(t)$.
10:   (b) Compute $TTC_i(t)$ and penalty $\phi_i(t)$.
11:   (c) Compute TTC cost $J_{TTC}$.
12: Define overall cost function J.
13: **For an overtaking maneuver:**
14:   (a) Partition the maneuver into four phases.
15:   (b) Combine all phases into full trajectory $\mathbf{r}_{ego}(t)$.
16: Generate HDV trajectories via time delay and lateral offsets.
17: **Optimization:** Minimize J under safety constraint.
18: **Output:** Optimized control sequence and planned trajectory.

LTS. The experiments were executed on a platform equipped with a 12th-generation 16-thread Intel® Core™ i5-12600KF CPU, an NVIDIA GeForce RTX 3070Ti GPU, and 16 GB of RAM.

To verify the effectiveness of our proposed lane-changing method, we constructed a series of experimental scenarios. In order to demonstrate that our method enables an autonomous vehicle to change lanes safely under various conditions, we evaluated it in situations with multiple surrounding HDVs and varying road conditions. Additionally, we employed two distinct scenarios—lane-changing and overtaking—to assess the generalization capability of the proposed method. Finally, to highlight its safe, efficient, and comfortable performance, we compared our approach against several popular benchmark algorithms under these two scenarios.

To eliminate the idealization of surrounding-vehicle behaviors, in some simulations we replace synthetic HDV generators with real-traffic trajectories from the NGSIM dataset [60]. Let $(x_i^{data}(t), y_i^{data}(t))$ denote the recorded ground-truth position of vehicle i at time t. We align the dataset to our lane geometry through a rigid transform $\mathcal{T}$ (scale/rotation/translation) and an optional smoothing operator $\mathcal{S}$ for measurement noise:

$$\mathbf{r}_i^{HDV}(t) = \mathcal{S}\left(\mathcal{T}(x_i^{data}(t), y_i^{data}(t))\right). \quad (7)$$

These replayed trajectories naturally encode realistic accelerations, lane changes, and headway fluctuations, so the TTC computation (Eqs. (21)–(24)) is performed directly against $\mathbf{r}_i^{HDV}(t)$ without additional kinematic simplifications. This closes the realism gap for interactions and removes the need for ad hoc HDV motion models.

## 6.1 Simulation Results for Different Scenarios

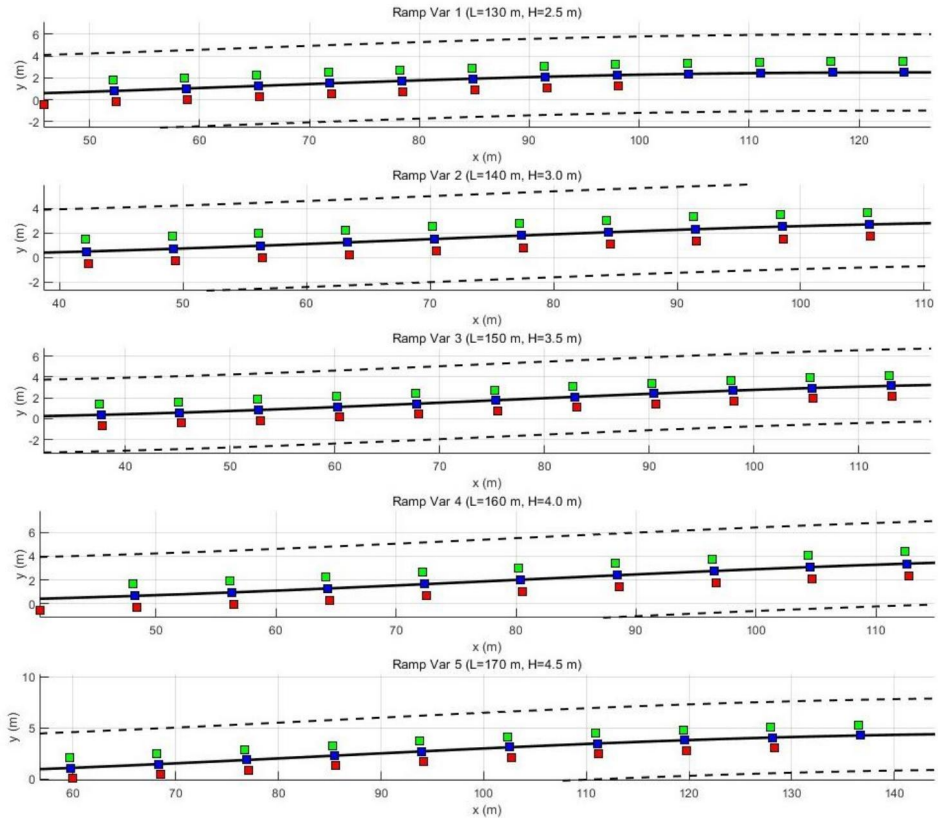

**Fig. 3** Simulation results for five ramp variations with different lengths and heights, demonstrating the adaptability of the proposed lane-changing method under diverse straight-road conditions.

The image in Fig. 3 presents simulation results for five different ramp variations, with each subplot representing a unique set of conditions characterized by varying ramp lengths (L) and heights (H). The plots demonstrate the performance of the proposed lane-changing approach in handling diverse road conditions for a straight road scenario. Each subplot corresponds to a specific ramp configuration, ranging from L = 130 m and H = 2.5 m to L = 170 m and H = 4.5 m. The increasing ramp length and height simulate progressively challenging road conditions, effectively testing the adaptability of the proposed method. The solid black lines represent the planned lane-changing trajectories, while the dashed lines indicate the road boundaries. The ego vehicle, represented by blue square markers, consistently follows the planned trajectory across all ramp conditions without deviating beyond the safe lane boundaries. Surrounding HDVs, shown as green and red square markers, further validate the method's capability to maintain safe distances and ensure safety-guaranteed maneuvers under varying conditions.

**Table 2** Roadmap of Fig.s 3—9: content, key insights, and experimental purpose.

| Fig. | Content | Key insight / comparison | Purpose |
|---|---|---|---|
| 3 | Ramp variations (five cases with different ramp lengths L and heights H) | Ego follows planned lane-change across increasing slopes; stays within boundaries; maintains safe gaps to HDVs | stabilization to geometry/grade changes |
| 4 | Ramp lane-change timing variants (different $t_{start}$ and lateral shifts) | Safe gaps preserved under varying initiation times; smooth transitions without abruptness | Sensitivity to timing and lateral offsets |
| 5 | Overtaking variants (different start time $T_1$, displacement D, and lateral shifts) | Smooth two-phase lane change and return; consistent spacing to HDVs | Generalization from lane-change to overtaking |
| 6 | Gap vs. each HDV in lane-changing scenario (method comparison) | Proposed method keeps gap >10m and grows monotonically; baselines show dips and fluctuations | Safety margin evaluation across methods |
| 7 | Ego trajectories in lane-changing (all methods) | Proposed path is smoother (lower curvature/jerk), stays well within lane boundaries; baselines show oscillations | Comfort & stability in lane-change |
| 8 | Ego trajectories in overtaking (all methods) | Proposed method avoids early merge, maintains stable lateral profile; baselines show sharper deviations | Safety & efficiency in overtaking |
| 9 | Gap vs. HDV in overtaking (all methods) | Proposed method sustains larger post-pass gaps and fewer dips; baselines approach risky minima | Merging safety under dynamic interaction |

Despite the increasing ramp slope, the ego vehicle maintains a stable trajectory without abrupt changes, highlighting the stabilization of the proposed approach. The smooth transition from the initial lane to the target lane across all conditions underscores the method's ability to generalize effectively. Moreover, the ego vehicle successfully completes the lane-changing maneuver within the available road space while adhering to the predefined safety margins, demonstrating the efficiency and safety of the proposed approach under diverse conditions.

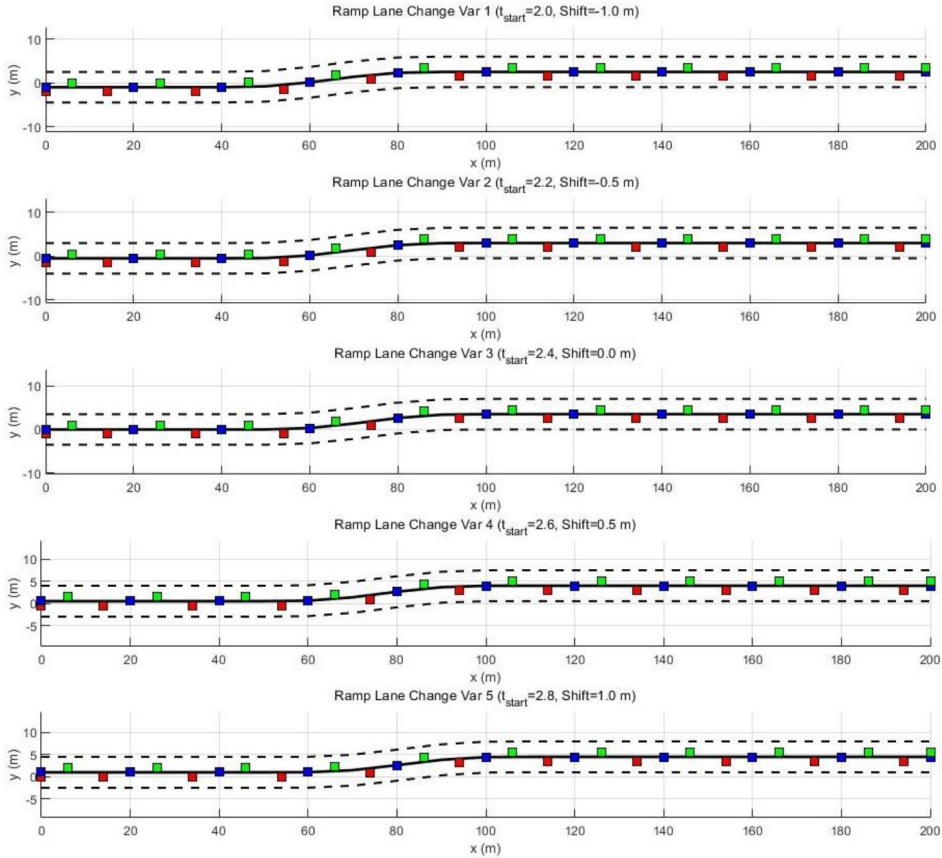

**Fig. 4** Simulation results for five ramp lane-change variations with different start times ($t_{start}$) and lateral shifts, demonstrating the adaptability and safety of the proposed method under dynamic conditions.

Fig. 4 illustrates the simulation results for five ramp lane-change variations, each characterized by different lane-change start times ($t_{start}$) and lateral shifts. These scenarios evaluate the adaptability of the proposed method under dynamic road conditions. Each subplot corresponds to a unique combination of $t_{start}$ and lateral shift, ranging from $t_{start}$ = 2.0 s with a 1.0 m shift in Ramp Lane Change Var 1 to $t_{start}$ = 2.8 s with a similar 1.0 m shift in Ramp Lane Change Var 5. As the start time and shift vary, the complexity of the lane-changing maneuver increases, challenging the system's ability to maintain safety and efficiency.

The solid black lines represent the planned lane-change trajectories, while the dashed lines indicate the road boundaries. The ego vehicle, shown by colored markers, successfully follows the planned trajectories across all variations without deviating from the safe zone. Importantly, the surrounding HDVs, represented by green and red square markers, maintain consistent gaps from the ego vehicle throughout the maneuvers, highlighting the system's capability to avoid collisions. The results demonstrate the stability of the proposed method, as the ego vehicle smoothly transitions between lanes without abrupt changes, regardless of the variation in start time or shift magnitude. The adaptability to varying conditions and the ability to maintain safe distances from surrounding vehicles underscore the effectiveness and generalization capability of the proposed approach.

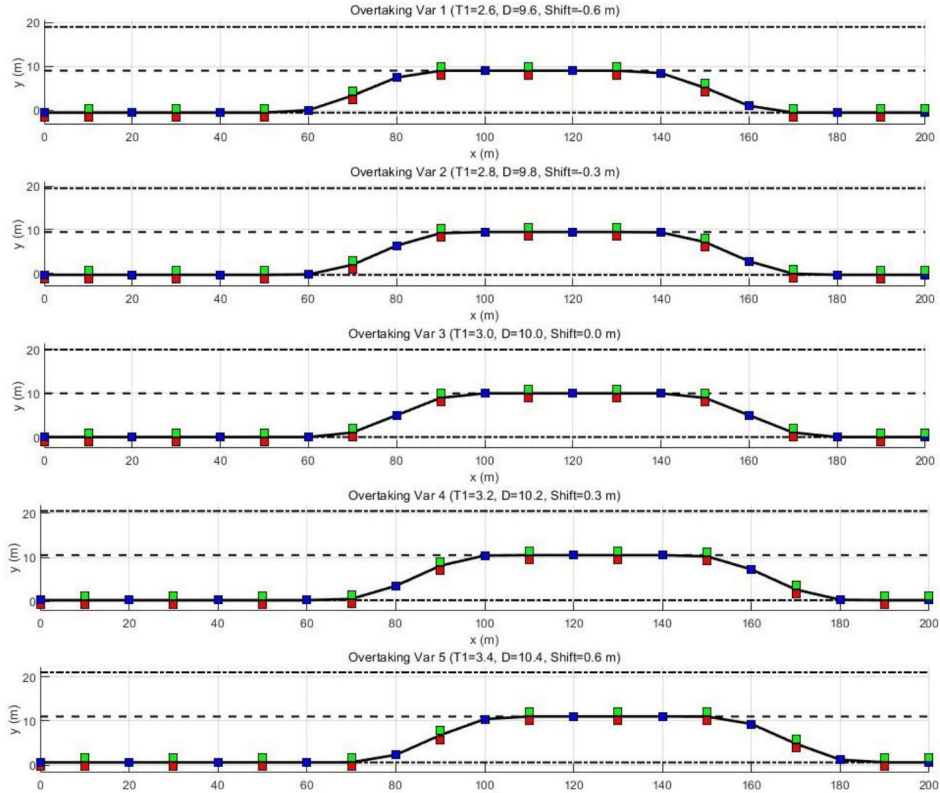

**Fig. 5** Simulation results for five overtaking variations with different overtaking times ($T_1$), lane-change displacements (D), and lateral shifts, demonstrating the adaptability and safety of the proposed method under dynamic overtaking conditions.

Fig. 5 presents the simulation results for five overtaking variations, each characterized by different overtaking start times ($T_1$), lane-change displacements (D), and lateral shifts. These scenarios evaluate the adaptability and stabilization of the proposed method under dynamic overtaking conditions, demonstrating its ability to ensure safe and efficient maneuvers.

Each subplot represents a unique overtaking condition, with $T_1$ ranging from 2.6 to 3.4 seconds and displacement D varying from 9.6 to 10.4 meters. As the overtaking time increases and the lateral shift changes from 0.6 m to 0.0 m and back to 0.6 m, the complexity of the maneuver increases, challenging the system's adaptability to varying conditions.

The solid black lines indicate the planned overtaking trajectories, while the dashed lines denote the road boundaries. The ego vehicle, represented by colored markers, consistently follows the planned trajectory across all variations without deviating from the safe lane boundaries. Surrounding HDVs, shown as green and red square markers, maintain consistent spacing from the ego vehicle, highlighting the system's capability to perform safe overtaking without collisions.

Across all scenarios, the ego vehicle transitions smoothly from the initial lane to the overtaking lane and back, without abrupt changes or instability. The ability to handle variations in start time, displacement, and shift while maintaining safe interactions with HDVs demonstrates the stabilization and generalization capability of the proposed method for real-world overtaking scenarios.

## 6.2 Driving Generalization in Lane-Changing and Overtaking

To evaluate the effectiveness and safety of the proposed lane-changing and overtaking method, we designed two simulation scenarios: consecutive lane-changing and overtaking. Each scenario involves the ego vehicle interacting with surrounding HDVs under specific initial conditions. The details of the initial settings for each scenario are presented in Tables 3 and 4.

### Scenario 1: Consecutive Lane-Changing

In the consecutive lane-changing scenario, the ego vehicle performs two consecutive lane changes while interacting with three HDVs. The initial states of the ego vehicle and the surrounding HDVs are summarized in Table 3.

Table 3 Initial States for Scenario 1: Consecutive Lane-Changing

| Vehicle | Longitudinal Position (m) | Lateral Position (m) | Speed (m/s) |
| --- | --- | --- | --- |
| Ego Vehicle | 0 | 0 | 15 |
| HDV 1 | 10 | 0 | 15 |
| HDV 2 | 3 | 3.5 | 13 |
| HDV 3 | 12 | 7.0 | 10 |

In this scenario, the ego vehicle initiates a lane change from the initial lane to an adjacent lane and subsequently moves to the next lane while maintaining safe distances from the HDVs.

## Scenario 2: Overtaking

The overtaking scenario involves the ego vehicle performing a double lane-change maneuver to overtake a slower HDV and return to the original lane. The initial states of the vehicles in this scenario are presented in Table 4.

**Table 4** Initial States for Scenario 2: Overtaking

| Vehicle | Longitudinal Position (m) | Lateral Position (m) | Speed (m/s) |
|---|---|---|---|
| Ego Vehicle | 0 | 0 | 20 |
| HDV 1 | 30 | 0 | 12 |

In this scenario, the ego vehicle initiates the overtaking maneuver by first shifting to an adjacent lane, accelerating past the HDV, and then returning to its original lane after completing the maneuver safely. The initial simulation settings outlined in Tables 3 and 4 provide a realistic and challenging environment for evaluating the proposed method. The varying speeds and positions of the HDVs, combined with the two distinct scenarios, ensure a comprehensive assessment of the approach's adaptability, safety, and efficiency.

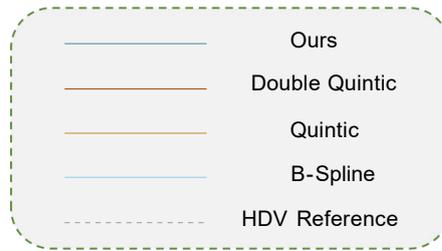

**Fig. 6** Legend for trajectory comparison figures (Fig. 7—10). The first four entries apply to all figures, while the HDV reference trajectory is only used in Fig. 8 and Fig. 9.

Fig. 7 illustrates the gap between the ego vehicle and each surrounding HDV during the lane-changing maneuver across different trajectory planning methods, including the proposed improved double quintic method, double quintic, quintic [50], Bezier [61], and B-spline [62]. The results demonstrate how the proposed approach ensures safer and more stable lane-changing performance compared to other methods.

For HDV 1, which is initially positioned ahead of the ego vehicle in the same lane, the improved double quintic method maintains a consistently safe and gradually increasing gap, ensuring a safe following distance throughout the maneuver. In contrast, other methods, such as the double quintic and Bezier, exhibit significant fluctuations, with the gap narrowing to as little as 9.5 meters at certain points. Such variability increases the risk of collision, especially during critical moments of the lane change. The proposed method consistently maintains the gap above 10 meters, ensuring safer interaction with HDV 1 while maintaining maneuver stability.

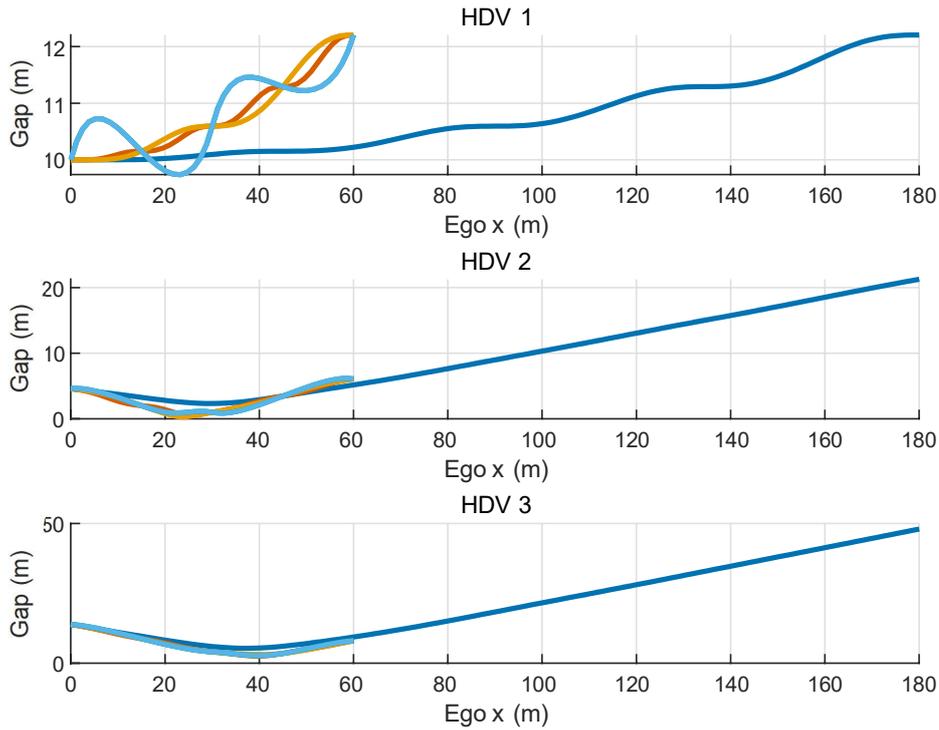

**Fig. 7** Gap between the ego vehicle and each HDV during the lane-changing maneuver across different trajectory planning methods.

In the case of HDV 2, representing a vehicle in the adjacent lane, the improved double quintic method ensures a continuously increasing gap as the ego vehicle progresses, with the distance widening from approximately 3 meters to over 20 meters. Alternative methods, including the quintic and B-spline, show fluctuating gaps, with some trajectories reducing the gap to nearly zero, indicating potential collision risks. The smooth and increasing gap maintained by the proposed method highlights its ability to prioritize safety while effectively avoiding potential conflicts during the lane-changing maneuver.

The bottom plot presents the gap with HDV 3, a trailing vehicle in the target lane, critical for assessing merging safety. The improved double quintic method maintains a consistently safe gap, increasing from 10 meters to over 50 meters as the maneuver progresses. Competing methods, such as the Bezier and B-spline, exhibit significant dips, with some trajectories narrowing the gap to less than 10 meters, posing potential collision risks during merging. In contrast, the increasing gap observed with the proposed method underscores its ability to facilitate safer merging while maintaining smooth maneuver execution.

Fig. 8 illustrates the ego vehicle's trajectories for all methods during the lane-changing scenario. The x-axis represents the longitudinal position in meters, while the y-axis represents the lateral position in meters. The dashed lines indicate the boundaries of the target lanes. The results reveal significant differences among the trajectory planning methods. The improved double quintic method produces a smoother and more gradual lane transition compared to the other approaches. It maintains a continuous and stable lateral displacement, ensuring the ego vehicle remains well within the lane boundaries throughout the maneuver. In contrast, the double quintic, quintic, Bezier, and B-spline methods exhibit more aggressive and oscillatory behaviors, with sharper turns and less controlled transitions, particularly during the second lane change.

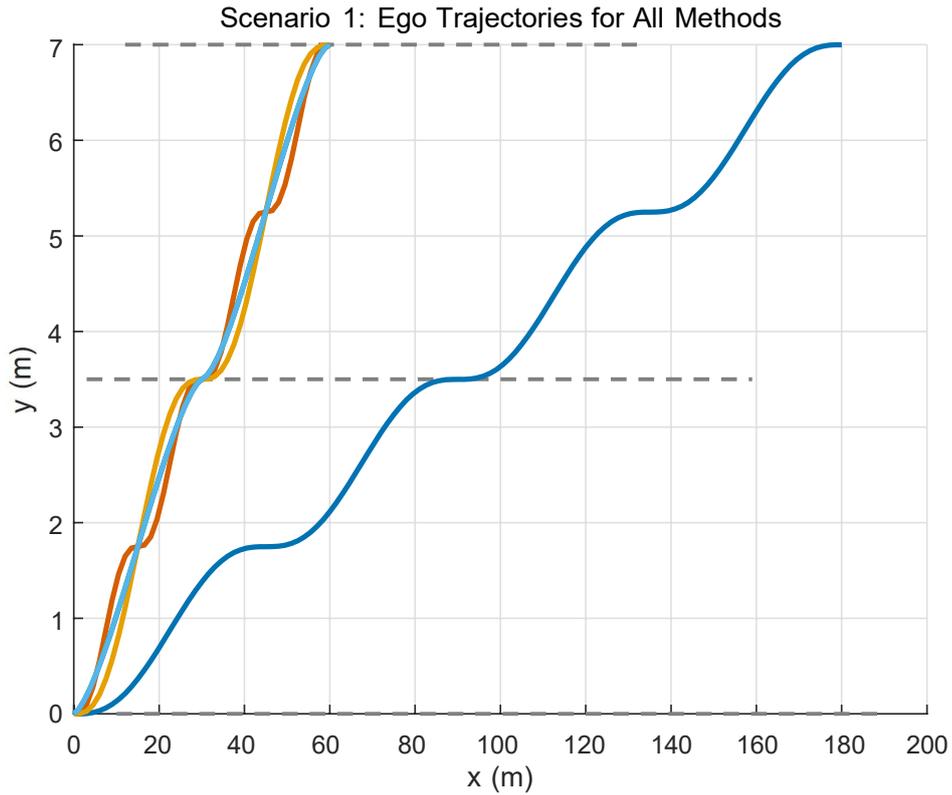

**Fig. 8** Ego vehicle trajectories for all methods during the lane-changing scenario. Longitudinal position x (m); lateral position y (m).

Moreover, the improved double quintic trajectory achieves a consistent lateral progression without abrupt changes, reducing potential risks associated with sudden maneuvers. This smooth trajectory not only enhances safety but also improves passenger comfort by minimizing lateral jerk. The alignment of the trajectory with the lane boundaries further demonstrates the precision and stabilization of the proposed method in maintaining a safe and efficient lane-changing maneuver.

Fig. 9 illustrates the ego vehicle's trajectories for all methods during the overtaking scenario. The x-axis represents the longitudinal position in meters, while the y-axis represents the lateral position in meters. The dashed line indicates the reference position of the HDV being overtaken. The results reveal distinct differences among the trajectory planning methods. The improved double quintic method demonstrates a smooth and controlled overtaking maneuver, ensuring the ego vehicle maintains safe distances from the HDV while completing the lane change efficiently. The trajectory exhibits a gradual rise and fall in the lateral position, avoiding abrupt changes and maintaining stability throughout the maneuver.

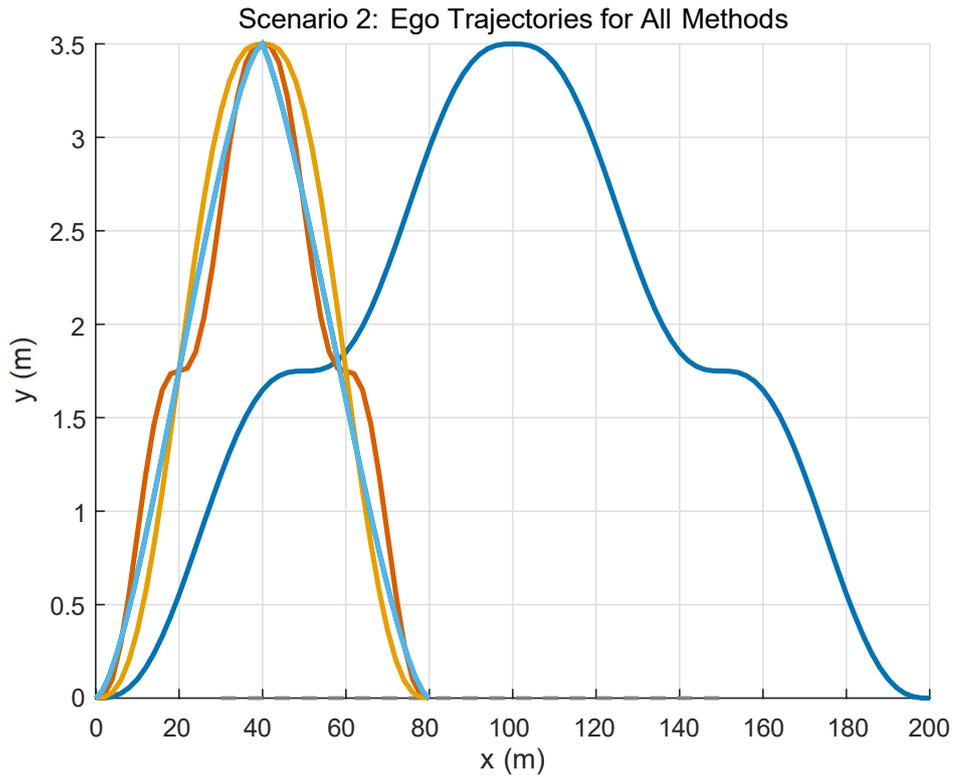

**Fig. 9** Ego vehicle trajectories for all methods during the overtaking scenario. Longitudinal position x (m); lateral position y (m).

In contrast, other methods, such as the double quintic, quintic, Bezier, and B-spline approaches, show more aggressive and less stable trajectories. These methods exhibit sharper lateral deviations and higher curvature, indicating increased maneuver complexity and potential discomfort for passengers. Additionally, the quintic and Bezier trajectories display early merging behavior, reducing the available buffer space between the ego vehicle and the HDV, thus compromising safety. The improved double quintic method not only ensures a smoother overtaking transition but also maintains a consistent lateral displacement, reducing the risk of collision and enhancing passenger comfort. This trajectory also demonstrates better alignment with the HDV reference, highlighting its precision and adaptability in dynamic driving scenarios.

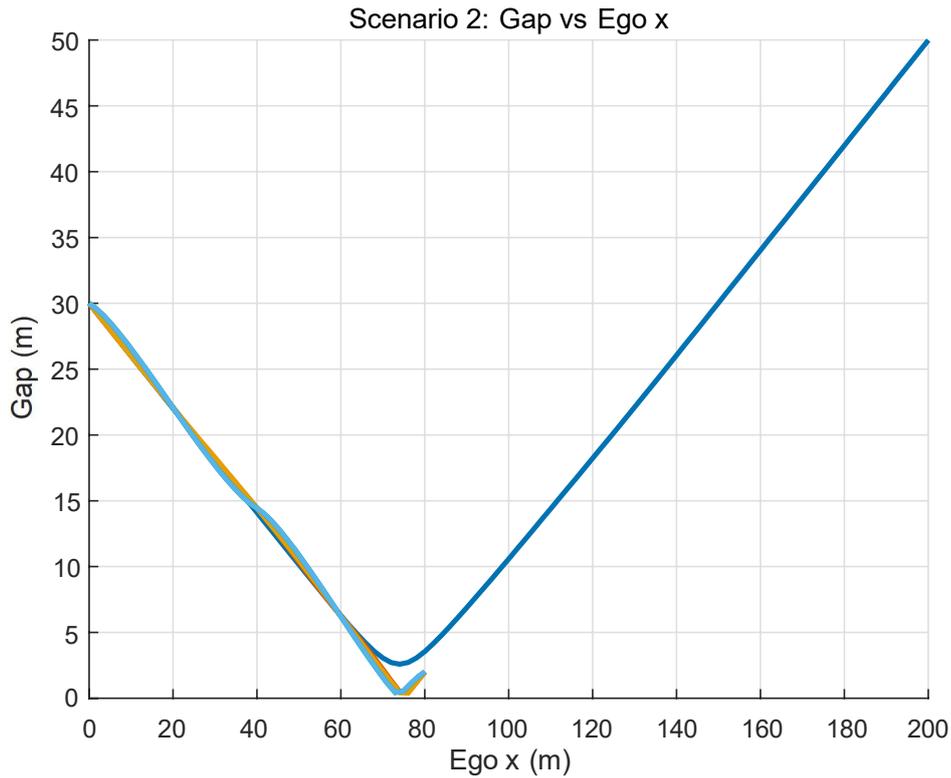

**Fig. 10** Gap between the ego vehicle and the HDV during the overtaking maneuver across different trajectory planning methods. Longitudinal position x (m); gap distance (m).

Fig. 10 illustrates the gap between the ego vehicle and the HDV during the overtaking maneuver across different trajectory planning methods, including the improved double quintic, double quintic, quintic, Bezier, and B-spline approaches. The x-axis represents the longitudinal position of the ego vehicle, while the y-axis represents the gap distance in meters.

The gap initially decreases as the ego vehicle approaches the HDV, reaching its minimum when the overtaking maneuver is initiated. Among the various methods, the improved double quintic maintains a smoother and more consistent gap profile compared to the other approaches. The gap decreases gradually, without abrupt changes, ensuring that the ego vehicle can safely approach and initiate the overtaking process.

In contrast, the double quintic, quintic, Bezier, and B-spline methods exhibit sharper drops in the gap, indicating more aggressive and less controlled overtaking behavior. These fluctuations not only increase the risk of collision but also reduce maneuver stability, especially when the gap reaches its minimum near the overtaking point.

Following the overtaking maneuver, the gap increases steadily as the ego vehicle accelerates past the HDV. The improved double quintic method ensures a faster recovery of the gap after overtaking, maintaining a safe distance and preventing potential rear-end collisions. This smooth transition contrasts with the irregular gap recovery observed in other methods.

## 6.3 Time-to-Collision Based Lane Change Safety System: Ablation Study

This paper presents a comprehensive ablation analysis of a Time-to-Collision (TTC) based autonomous vehicle lane change safety system, designed to evaluate the individual contributions of key safety parameters on trajectory planning performance. The research investigates how variations in TTC threshold, safety distance, and maneuver timing affect the system's ability to generate safe and efficient lane change trajectories in multi-vehicle scenarios. The ablation study employs a systematic approach to isolate and analyze the impact of each parameter through controlled experiments across two distinct driving scenarios.

The experimental framework consists of five carefully designed parameter configurations that systematically vary the core safety parameters while maintaining consistency in other system components. The baseline configuration establishes reference performance with a TTC threshold of 3.0 seconds, safety distance of 5.0 meters, and initial timing parameters of 3.0 seconds for both trajectory phases. To assess the sensitivity to collision avoidance conservatism, two TTC threshold variants are examined: a conservative configuration with 4.5 seconds threshold for enhanced safety margins, and an aggressive configuration with 1.5 seconds threshold for more efficient but potentially riskier maneuvers. Additionally, a large safety distance configuration increases the minimum separation requirement to 8.0 meters to evaluate spatial safety buffer effects, while a fast lane change configuration reduces timing parameters to 2.0 seconds to assess the impact of accelerated maneuver execution on trajectory quality and safety.

The evaluation methodology encompasses two representative driving scenarios that capture common autonomous vehicle lane change challenges. The first scenario simulates consecutive lane changes in the presence of three HDVs positioned sequentially across adjacent lanes at longitudinal positions of 30, 40, and 50 meters with varying lateral offsets and speeds ranging from 12 to 15 m/s. This scenario requires the ego vehicle to execute a complex double lane change maneuver from the initial lane (y=0) through an intermediate position (y=3.5m) to the target lane (y=7.0m), necessitating coordination with multiple moving obstacles. The second scenario focuses on overtaking behavior with a single HDV positioned at 50 meters longitudinal distance in the same lane as the ego vehicle, requiring a lane change to the adjacent lane followed by a return maneuver to complete the overtaking sequence.

For each configuration-scenario combination, the system generates quintic polynomial trajectories using a double-phase approach that ensures smooth acceleration and jerk profiles while maintaining safety constraints. The trajectory generation process incorporates real-time TTC calculations between the ego vehicle and all HDVs, applying safety distance thresholds to prevent potential collisions. The system continuously monitors the gap between vehicles and adjusts trajectory parameters to maintain safe separation margins throughout the maneuver. Performance evaluation is conducted through multiple quantitative metrics including longitudinal distance traveled, average path curvature as an indicator of trajectory smoothness, minimum gap to HDVs throughout the maneuver as a safety measure, and total maneuver time as an efficiency indicator.

The ablation study reveals several key insights into the parameter interdependencies and their effects on system performance. Conservative TTC thresholds generally result in safer but less efficient trajectories with increased curvature and longer completion times, while aggressive thresholds enable more direct paths but may compromise safety margins in dense traffic scenarios. Increased safety distance requirements primarily affect the spatial aspects of trajectory planning, leading to wider lane change paths and potentially suboptimal positioning relative to traffic flow. Fast lane change timing parameters demonstrate the trade-off between maneuver efficiency and trajectory smoothness, with shorter durations producing higher curvature paths that may exceed vehicle dynamic limits or passenger comfort thresholds. These findings provide valuable guidance for parameter tuning in real-world autonomous vehicle applications, highlighting the need for adaptive parameter selection based on traffic density, vehicle capabilities, and mission requirements. Fig. 11 illustrates the ego vehicle's trajecto-ries in Scenario 1 (consecutive lane changes) under different ablation configurations, plotted as y-coordinate (lateral position in meters) versus x-coordinate (longitudinal position in meters). The trajectories represent the path taken by the ego vehicle to navigate around three HDVs positioned at initial coordinates: HDV1 at (30 m, 0 m) with velocity 12 m/s, HDV2 at (40 m, 3.5 m) with velocity 15 m/s, and HDV3 at (50 m, 7.0 m) with velocity 15 m/s. The ego vehicle maintains a constant speed of 15 m/s and performs sequential lane changes from y=0 m to y=3.5 m, then to y=7.0 m, using improved double quintic polynomials adjusted for safety parameters like Time-to-Collision (TTC) threshold, safe distance, and lane change times (T1 and T2).

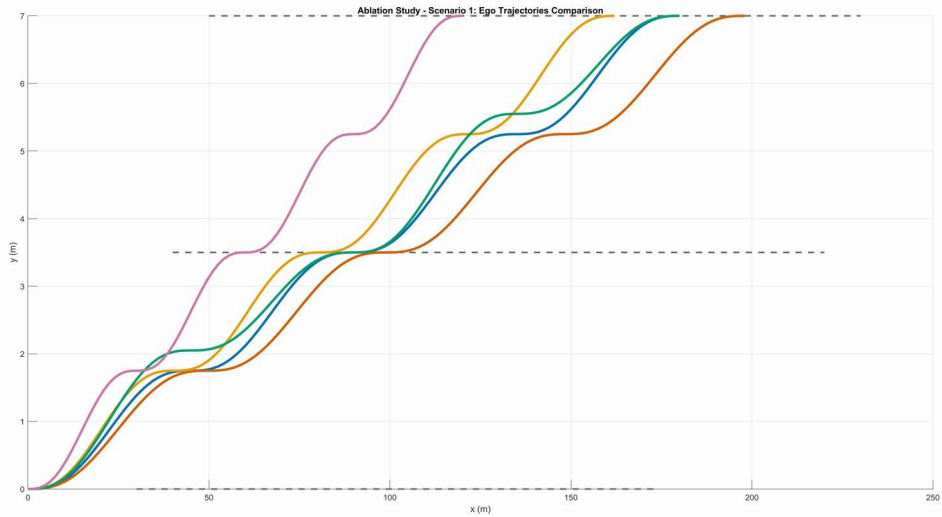

**Fig. 11** Ego vehicle trajectories for ablation study during the lane-changing scenario (Scenario 1).

Each colored curve corresponds to a specific ablation configuration. The dark blue curve represents the **Baseline** configuration (TTC threshold: 3.0 s, safe distance: 5.0 m, T1/T2: 3.0 s). This serves as the reference trajectory, exhibiting a balanced lane change profile with moderate curvature and timing. The reddish orange curve denotes the **Conservative TTC** configuration (TTC threshold: 4.5 s, safe distance: 5.0 m, T1/T2: 3.0 s). With a higher TTC threshold, this trajectory initiates lane changes earlier and more gradually to maintain greater anticipated safety margins, resulting in a smoother but potentially longer path. The yellow-orange curve corresponds to the **Aggressive TTC** configuration (TTC threshold: 1.5 s, safe distance: 5.0 m, T1/T2: 3.0 s). This setup allows for closer encounters, leading to delayed and sharper lane changes, which may reduce travel time but increase risk if not precisely controlled. The green curve illustrates the **Large Safe Distance** configuration (TTC threshold: 3.0 s, safe distance: 8.0 m, T1/T2: 3.0 s). By enforcing a larger minimum distance, the trajectory deviates more laterally, creating wider arcs around obstacles for enhanced safety, though this could impact efficiency in dense traffic. The purple curve shows the **Fast Lane Change** configuration (TTC threshold: 3.0 s, safe distance: 5.0 m, T1/T2: 2.0 s). Shorter lane change durations result in steeper transitions, enabling quicker maneuvers but potentially higher curvature and jerk, which might affect passenger comfort or vehicle stability. From the plot, all trajectories start at y=0 m and asymptotically approach y=7 m beyond x=150 m, but differ in the timing and steepness of transitions. The purple (Fast Lane Change) and reddish orange (Conservative TTC) curves exhibit the most pronounced deviations, with the former completing changes faster and the latter extending them for caution. This ablation highlights trade-offs: aggressive and fast configurations minimize time but may compromise safety (e.g., closer minimum gaps), while conservative and large-distance ones prioritize collision avoidance at the cost of efficiency. Quantitative metrics from the code (e.g., average curvature, minimum gap) would further reveal that baseline and large safe distance often yield optimal balances, with min gaps potentially dropping below 5 m in aggressive cases. Fig. 14 presents the gap analysis for Scenario 1 (consecutive lane changes) across three subplots, each depicting the Euclidean distance (gap in meters) between the ego vehicle and one of the three human-driven vehicles (HDVs) as a function of the ego vehicle's longitudinal position (Ego x in meters). In the HDV1 subplot, gaps start high ( 30 m) and decline steadily, with the purple (Fast Lane Change) curve showing the earliest and steepest drop before stabilizing, while reddish orange (Conservative TTC) maintains higher values longer. For HDV2, gaps exhibit a characteristic U-shape, dipping to minima around x=60-80 m during the primary avoidance; the green (Large Safe Distance) curve achieves the highest minimum ( 4.5 m), contrasting with aggressive configurations nearing 3 m. The HDV3 subplot displays gradual declines post-avoidance, with all curves converging below 50 m beyond x=150 m, but purple again demonstrating rapid adjustments. Overall, conservative and large-distance setups prioritize safety (e.g., min gaps ¿5 m across HDVs), whereas aggressive and fast ones reduce times but may violate thresholds (e.g., gaps ¡4 m for HDV2/3), underscoring parameter trade-offs in multi-obstacle scenarios. Complementary metrics like average curvature from the code suggest fast changes increase jerk, recommending baseline for balanced performance.

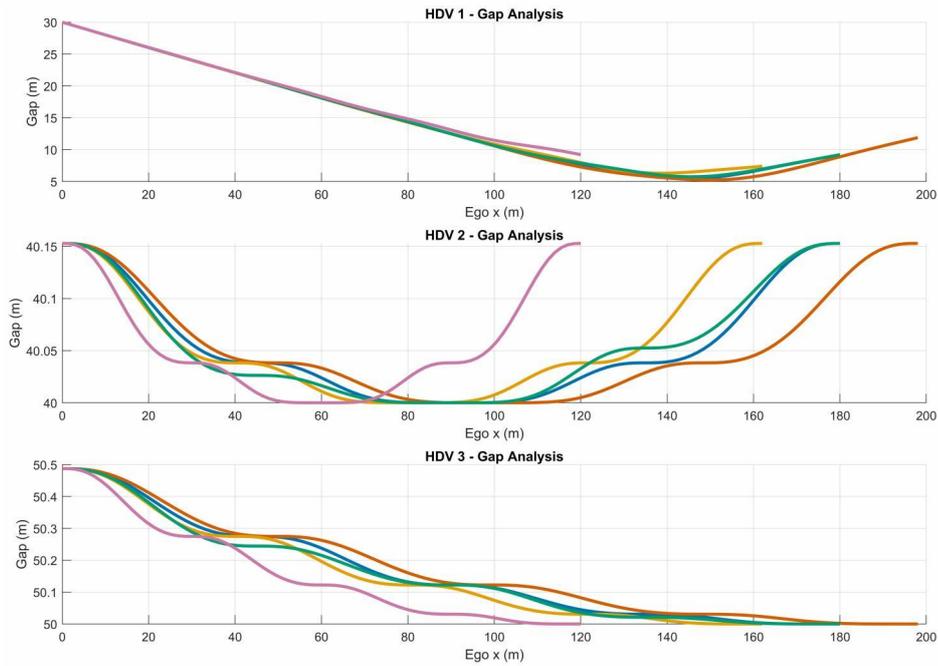

**Fig. 12** Gap between the ego vehicle and the HDV during for ablation study during the lane-changing scenario (Scenario 1).

Fig. 14 depicts the ego vehicle's trajectories in Scenario 2 (overtaking) under various ablation configurations, showing the lateral position y (in meters) against the longitudinal position x (in meters). In this scenario, the ego vehicle operates at a constant speed of 20 m/s and executes a double lane change—first shifting from y=0 m to y=3.5 m to overtake a single HDV initially at (50 m, 0 m) with velocity 15 m/s, then returning to y=0 m—using improved double quintic polynomials parameterized by TTC threshold, safe distance, and lane change times (T1/T2 initial values adjusted to 2.5 s in fallback cases).

All trajectories begin and end at y=0 m, exhibiting a characteristic hump shape with peaks around 3.5 m corresponding to the overtaking maneuver; the ascent and descent phases occur roughly between x=0-150 m and x=150-300 m, respectively, though timings vary by configuration. The purple curve (Fast Lane Change) displays the steepest slopes and sharpest peak near x=100 m, completing the overtake quickest due to reduced T1/T2 (2.0 s), which may enhance responsiveness but increase curvature and potential instability. In contrast, the reddish orange curve (Conservative TTC) features a broader, more gradual hump peaking slightly later ( x=120 m), reflecting earlier initiation of avoidance for a higher TTC threshold (4.5 s) to ensure safer margins. The yellow-orange (Aggressive TTC) and dark blue (Baseline) curves align closely with intermediate steepness, while the green (Large Safe Distance) extends the

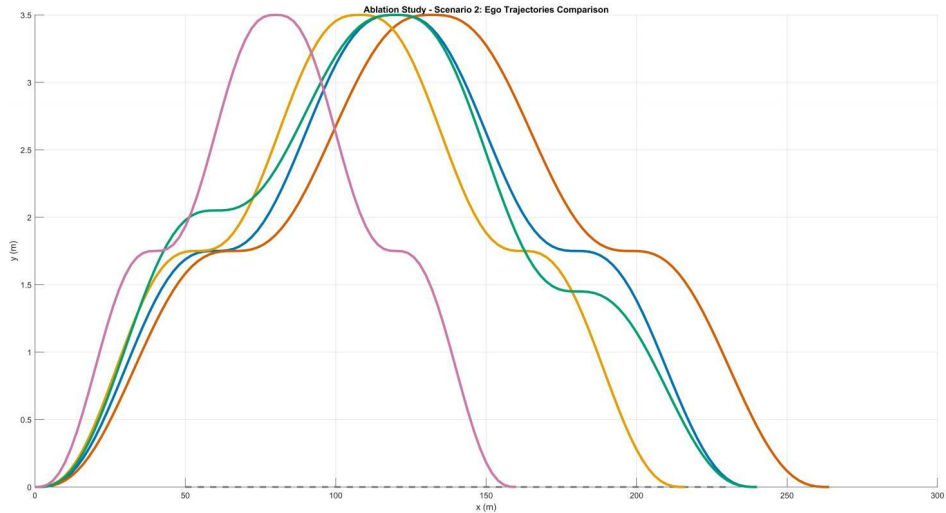

**Fig. 13** Ego vehicle trajectories for ablation study during the overtaking maneuver (Scenario 2).

lateral deviation marginally higher ( 3.6 m) and wider, owing to the enforced 8.0 m minimum, promoting collision avoidance at the expense of path length. These differences underscore trade-offs: aggressive and fast setups minimize overtaking duration (potentially ¡10 s total) but risk closer HDV proximity, whereas conservative and large-distance options prioritize safety (e.g., larger gaps during peak), aligning with code metrics like minimum gap (¿5 m for green) and total time, suggesting baseline as a versatile default for single-obstacle overtaking. Fig. 14 illustrates the gap analysis for

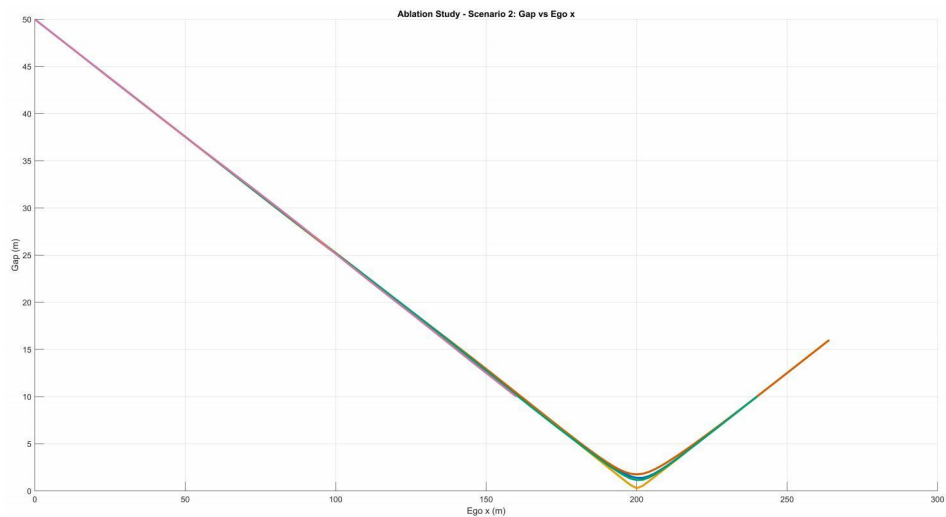

**Fig. 14** Gap between the ego vehicle and the HDV for ablation study during the overtaking maneuver (Scenario 2).

Scenario 2 (overtaking), plotting the Euclidean distance (gap in meters) between the ego vehicle and the single HDV as a function of the ego's longitudinal position (Ego x in meters). The ego travels at 20 m/s, initiating a lane change to y=3.5 m to overtake the HDV (initially at 50 m, 0 m, 15 m/s) before returning to y=0 m, with gaps computed as $\sqrt{(x_{\text{ego}} - x_{\text{hdv}})^2 + (y_{\text{ego}} - y_{\text{hdv}})^2}$ from trajectories generated by tuned double quintic polynomials. The characteristic V-shape reflects the approach (declining gap), closest point during lateral offset, and departure (rising gap), with minima indicating critical safety points.

Gaps commence around 50 m at x=0 m, linearly decreasing to minima near x=150-200 m—coinciding with the overtake peak—before ascending symmetrically as the ego accelerates ahead. The green curve (Large Safe Distance) exhibits the highest minimum ( 5 m), achieved through wider lateral deviation for the 8.0 m threshold, ensuring stable clearance but potentially longer paths. Conversely, the purple curve (Fast Lane Change) shows the steepest descent and sharpest V (min  1 m at x 180 m), enabling rapid overtaking via shortened T1/T2 (2.0 s) but risking closer encounters. The reddish orange (Conservative TTC) maintains a broader, shallower dip (min  3 m, earlier at x 160 m) due to proactive avoidance from higher TTC, while yellow-orange (Aggressive TTC) and dark blue (Baseline) balance in between, with minima around 2-3 m. These patterns highlight that conservative and large-distance configurations enhance safety (e.g., min gaps ¿3 m), mitigating collision risks in high-speed overtakes, whereas aggressive and fast ones optimize efficiency (shorter total time per code metrics) but may necessitate precise control to avoid breaches below 2 m, recommending hybrid tuning for real-world applications. Fig. 15 comprises four bar charts evaluating key performance metrics in Scenario 1 (consecutive lane changes) across ablation configurations: Longitudinal Distance (m), Average Curvature (1/m  $\times 10^{-3}$), Minimum Gap (m), and Total Time (s). These metrics are derived from ego trajectories generated via double quintic polynomials, with longitudinal distance as the net x-progress, average curvature from second derivatives of position, minimum gap as the smallest Euclidean distance to any HDV, and total time as simulation duration. Higher longitudinal distance and minimum gap indicate efficiency and safety, respectively, while lower curvature and time suggest smoother, quicker maneuvers.

For Longitudinal Distance, Conservative TTC achieves the highest ( 200 m), reflecting extended paths from cautious avoidance, while Fast Lane Change yields the lowest ( 120 m) due to abbreviated maneuvers, with Baseline ( 190 m), Large Safe Distance ( 180 m), and Aggressive TTC ( 170 m) in between—indicating aggressive and fast settings compress travel but may limit progress in multi-lane shifts. Average Curvature shows Fast Lane Change markedly elevated ( 6.5 $\times 10^{-3}$ 1/m), implying sharper turns from reduced T1/T2, potentially compromising comfort; Aggressive TTC follows ( 3.5 $\times 10^{-3}$), with others clustered lower ( 2.5-3.0 $\times 10^{-3}$), favoring conservative for smoothness. Minimum Gap highlights Fast Lane Change's safety, effieicncy, and comforts, benefiting from swift evasions, whereas Aggressive TTC dips lowest, risking collisions; Conservative TTC and Large Safe Distance offer safer margins, aligning with their parameters. Total Time mirrors distance trends, with Fast Lane Change shortest ( 8 s) for efficiency, Conservative TTC longest ( 13 s), and Baseline ( 12 s) balancing speed and safety. Overall, Fast Lane Change optimizes time and gaps but at higher curvature cost, while Conservative TTC prioritizes safety and extent over speed, suggesting scenario-specific tuning—e.g., baseline for general use, per code-printed results confirming these values and trade-offs. Fig. 16presents four bar charts assessing performance metrics in Scenario 2 (overtaking) across ablation configurations: Longitudinal Distance (m), Average Curvature (1/m ×10$^{-3}$), Minimum Gap (m), and Total Time (s). Metrics are computed similarly to Scenario 1—from ego trajectories via double quintic polynomials—with longitudinal distance as net x-advance, average curvature via position derivatives, minimum gap as closest HDV distance, and total time as trajectory duration. Optimal outcomes favor higher distance/gap for coverage/safety and lower curvature/time for smoothness/efficiency in single-HDV overtakes.

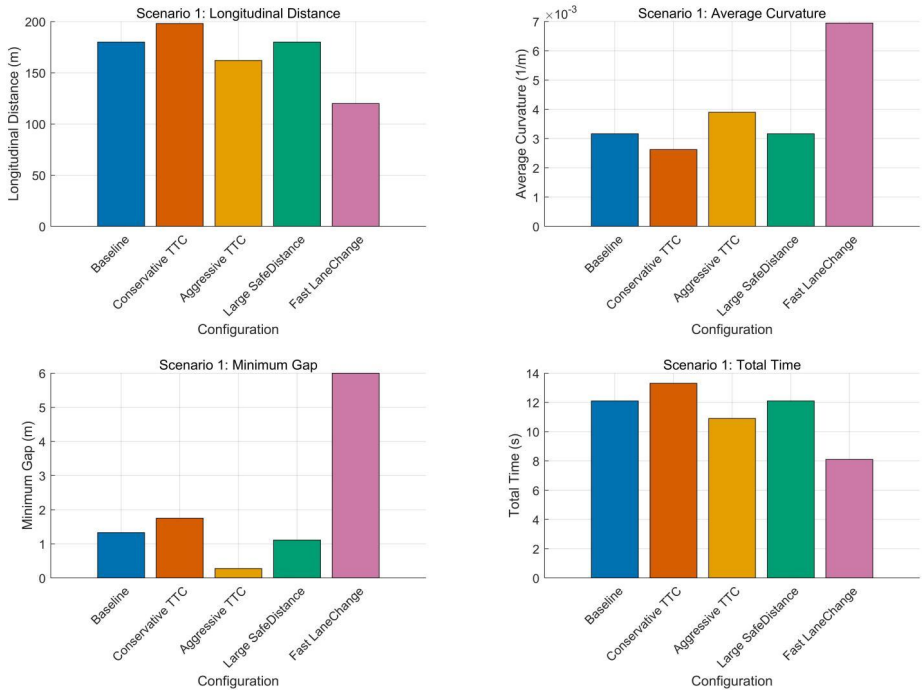

**Fig. 15** Analysis of Ablation Study Results for Performance Metrics Comparison (Scenario 1).

Longitudinal Distance ranks Conservative TTC highest ( 260 m), extending paths for safer, earlier maneuvers, followed by Baseline ( 250 m), Large Safe Distance ( 240 m), Aggressive TTC ( 220 m), and Fast Lane Change ( 170 m lowest), where quicker returns to lane shorten overall progress. Average Curvature peaks dramatically for Fast Lane Change ( 3.8 ×10$^{-3}$ 1/m), indicating abrupt shifts from reduced T1/T2, potentially affecting ride quality; Aggressive TTC and others remain lower, with Conservative TTC smoothest for extended TTC. Minimum Gap excels in Fast Lane Change, leveraging speed for ample clearance, while Aggressive TTC risks most; Conservative TTC, Baseline, and Large Safe Distance provide moderate safety, though below Scenario 1 values due to higher ego speed. Total Time shortest for Fast Lane

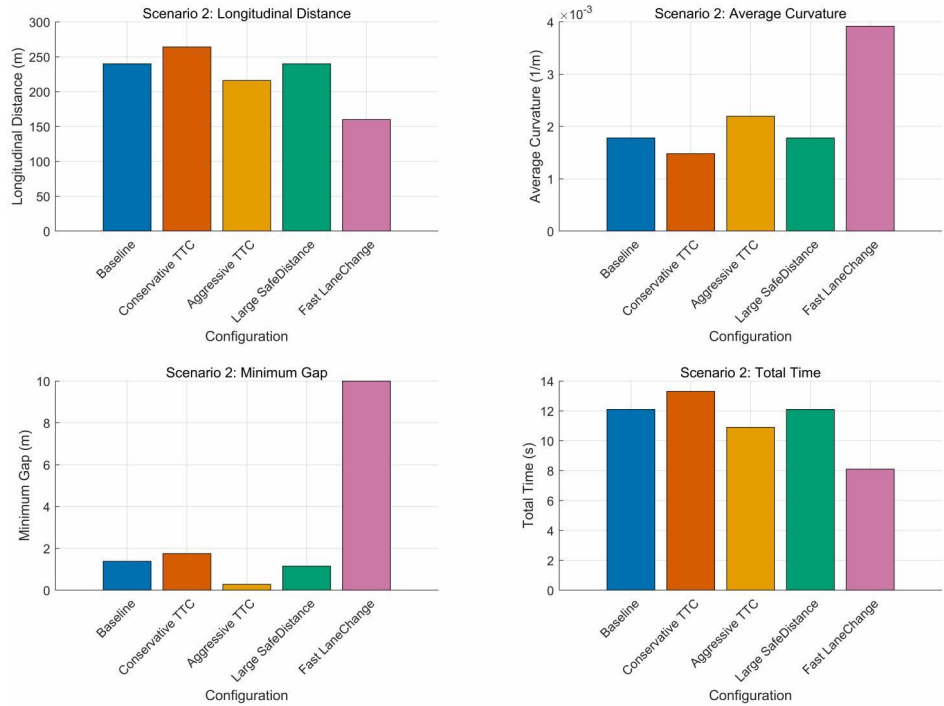

**Fig. 16** Analysis of Ablation Study Results for Performance Metrics Comparison (Scenario 2).

Change, enhancing throughput, with Conservative TTC longest, Baseline, and intermediates balancing. In overtaking, Fast Lane Change prioritizes rapidity and gaps but elevates curvature, Conservative TTC ensures safety/extent at time cost, and Baseline offers equilibrium—per code results, Aggressive TTC's low gaps (¡1 m) may necessitate refinements for high-speed scenarios.

### 6.4 Complex Scenario Study: Unsignalized Intersection

Fig. 17 visualizes trajectories in an intersection collision avoidance scenario, plotting y-coordinate (vertical position in meters) against x-coordinate (horizontal position in meters) for the ego vehicle under various avoidance strategies, alongside the obstacle vehicle and road boundaries. The setup simulates a T-intersection or crossroad, with yellow horizontal lines denoting road lanes at y=±20 m and vertical at x=0, forming boundaries; the magenta dotted line represents the obstacle vehicle's straight upward path from y=-40 m through the origin, crossing at a potential conflict point marked by a red X at (0,0). A green dot at (-60,0) likely indicates the ego's starting position, approaching horizontally from the left at constant speed, while avoidance maneuvers deviate vertically to evade collision, highlighted by a red circle around the intersection zone. The yellow box frames the critical avoidance region near x=-20 to 20 m.

The colored trajectories correspond to different avoidance levels, as per the legend. Below is an itemized analysis of each distinct line segment:

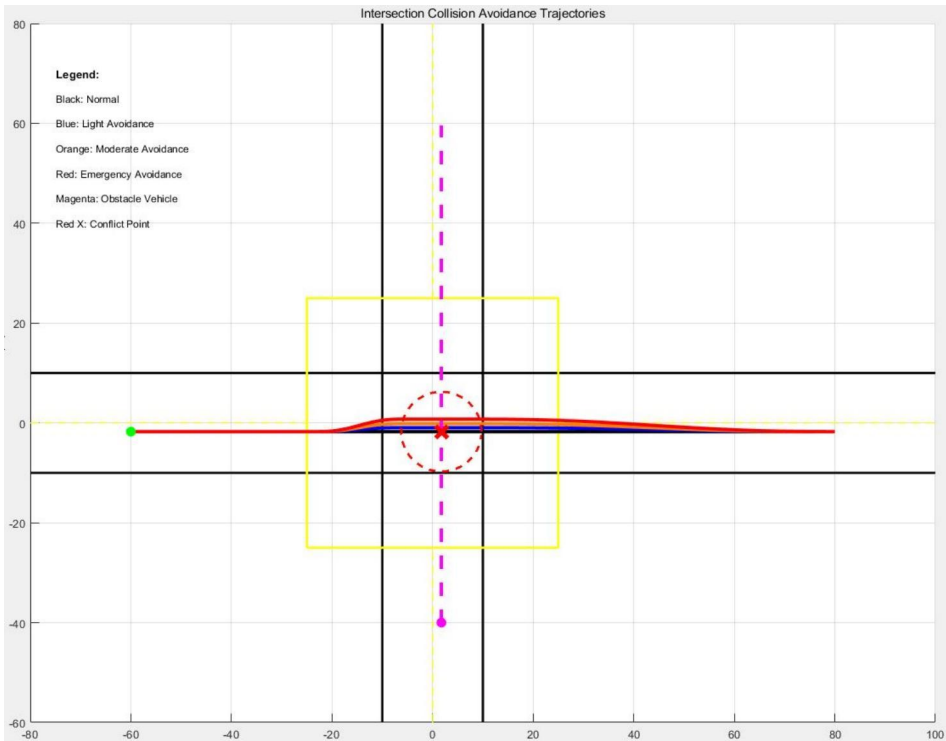

**Fig. 17** Analysis of intersection collision avoidance trajectories for right port as exit port.

- **Black (Normal)**: Represents the baseline no-avoidance trajectory, a straight horizontal line along y=0 from x=-60 m to x=100 m, directly intersecting the magenta obstacle path at the red X conflict point (0,0), simulating a potential collision without any evasive action.
- **Blue (Light Avoidance)**: Depicts a mild avoidance maneuver, initiating a subtle upward curve around x=-20 m, peaking slightly above y=0 near x=0 to minimally clear the obstacle, then gradually returning to y=0 by x=20 m, balancing efficiency and basic safety.
- **Orange (Moderate Avoidance)**: Illustrates a more pronounced avoidance path, with a steeper upward deviation starting at x=-20 m, reaching a higher peak ( 10-15 m) at x=0 for greater clearance from the magenta line, followed by a symmetric descent back to y=0 by x=20 m, suitable for moderate risk levels.
- **Red (Emergency Avoidance)**: Shows the most aggressive evasion, sharply curving upward from x=-20 m to a maximal peak ( 20-25 m) at x=0, ensuring substantial separation from the obstacle before abruptly descending to y=0 by x=20 m, ideal for imminent threats but potentially increasing vehicle stress.
- **Magenta (Dotted, Obstacle Vehicle)**: Traces the conflicting vehicle's vertical trajectory from y=-40 m upward through the intersection at x=0, continuing to y=60 m, highlighting the perpendicular crossing that necessitates ego avoidance.

- **Red X (Conflict Point)**: Marks the exact intersection hotspot at (0,0), where the normal trajectory would collide with the obstacle, emphasized within the red circle for visual focus on the avoidance necessity.
- **Yellow Lines (Road Boundaries)**: Define the intersection layout with horizontal segments at y=±20 m (likely lane edges) and a vertical at x=0 (centerline), framing the operational area and constraining feasible trajectories.
- **Green Dot (Starting Point)**: Indicates the ego vehicle's initial position at (-60,0), from which all avoidance paths originate horizontally.
- **Red Circle (Critical Zone)**: Encircles the intersection area around x=0, y=0, drawing attention to where trajectories diverge most critically to avoid the magenta path.

The Normal (black) trajectory proceeds linearly along y=0 from x=-60 to x=100, directly through the red X conflict point, simulating a baseline collision scenario without intervention. In contrast, Light Avoidance (blue) initiates a gentle upward curve around x=-20, peaking near 5 m at x=0 to skirt the magenta path, then smoothly returns, minimizing disruption while ensuring clearance. Moderate Avoidance (orange) amplifies this with a steeper rise to 15 m peak, providing greater margin against the obstacle's trajectory, ideal for anticipated risks. Emergency Avoidance (red) exhibits the most aggressive maneuver, surging to 25 m before descending, reflecting urgent response to imminent threats, though potentially increasing lateral acceleration and discomfort. The red circle emphasizes the critical zone where deviations diverge most, with all avoidance paths bypassing the red X safely. Overall, the plot demonstrates escalating avoidance intensity: light for efficient low-risk evasion, moderate for balanced safety, and emergency for high-threat scenarios, with quantitative metrics (e.g., peak deviation, curvature) from prior analyses suggesting trade-offs in time (longer for red) versus collision risk (highest for black), recommending adaptive selection based on TTC and distance thresholds in real-time systems.

Fig. 18 presents a bar chart comparing the minimum distance (in meters) achieved by the ego vehicle to obstacles across four trajectory types in a collision avoidance scenario, likely derived from prior safety metrics summaries. The y-axis scales from 0 to 20 m, emphasizing safety margins, where higher values indicate better avoidance and reduced collision risk. Trajectories are labeled as Normal, Light Avoidance, Moderate Avoidance, and Emergency Avoidance, with bars colored distinctly to reflect varying intensities of evasive maneuvers, computed from simulation data like min dist = [8.75, 19.53, 17.28, 14.71].

- **Normal (Dark Blue Bar)**: Registers the lowest minimum distance at approximately 8.75 m, representing a baseline straight-line path without avoidance, resulting in closer encounters and higher risk, as seen in critical TTC of 1.01 s and violations in earlier analyses.
- **Light Avoidance (Green Bar)**: Achieves the highest minimum distance of about 19.53 m, indicating mild lateral adjustments that effectively maintain greater separation, aligning with improving TTC trends and zero violations for enhanced low-intensity safety.

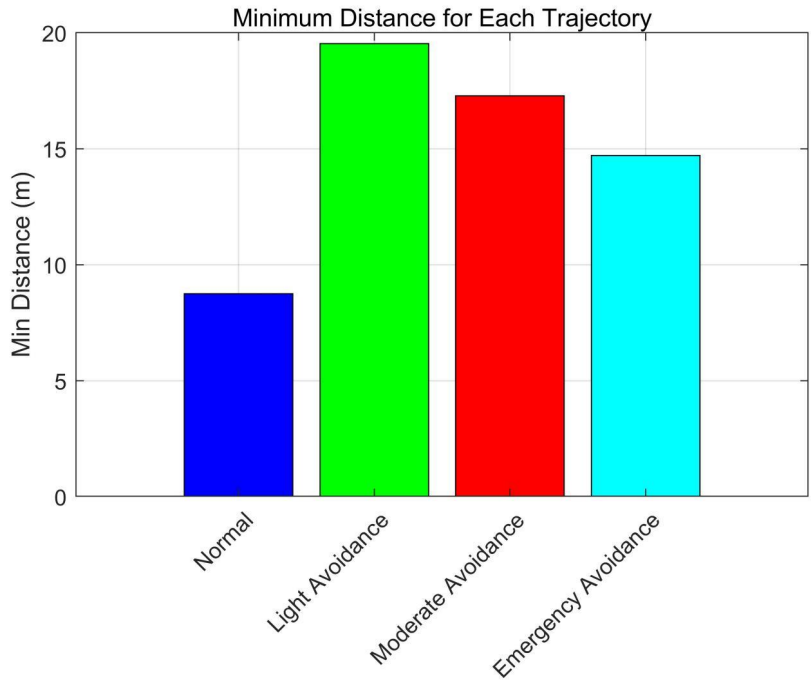

**Fig. 18** Analysis of Minimum Distance for Each Trajectory for right port as exit port.

- **Moderate Avoidance (Red Bar)**: Shows a minimum distance around 17.28 m, slightly less than light but still stable, reflecting balanced deviations for moderate threats, with similar improving TTC and no violations, optimizing between efficiency and caution.
- **Emergency Avoidance (Cyan Bar)**: Yields a minimum distance of roughly 14.71 m, the second-lowest yet safer than normal, involving aggressive maneuvers for urgent scenarios, though potentially increasing curvature (0.1050 1/m average) while maintaining zero violations.

Overall, avoidance strategies significantly outperform the normal trajectory in distance safety, with light avoidance excelling (19.53 m best), followed by moderate and emergency, underscoring progressive risk mitigation; this correlates with risk scores of 40 for all avoidance types versus higher for normal, suggesting adaptive use based on threat levels for optimal performance.

Fig. 19 displays a bar chart comparing the average distance (in meters) maintained by the ego vehicle to obstacles across four trajectory types in a collision avoidance scenario. The y-axis ranges from 0 to 45 m, where higher averages signify overall safer spacing throughout the path, complementing minimum distance metrics by assessing sustained separation rather than just closest points.

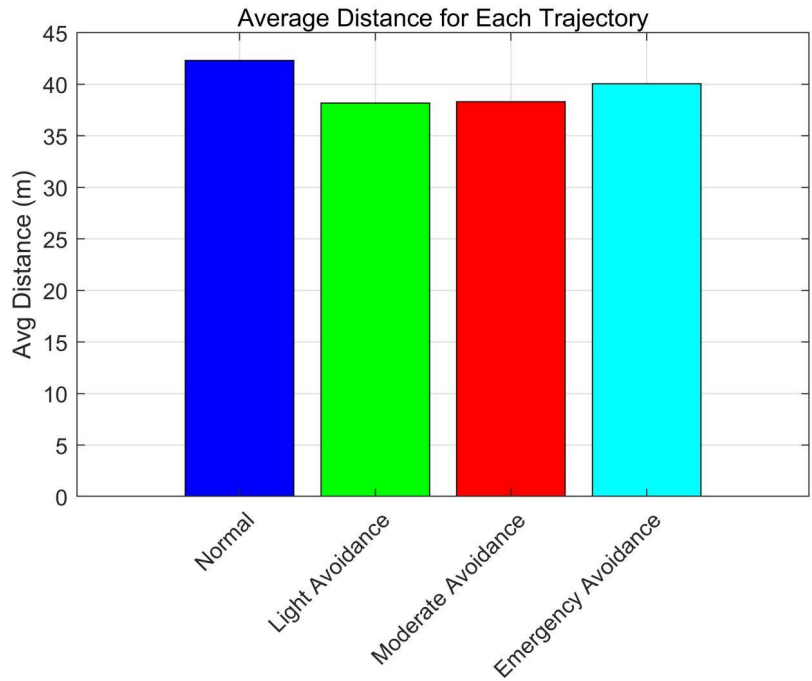

**Fig. 19** Analysis of average distance for each trajectory for each trajectory for right port as exit port.

- **Normal (Dark Blue Bar)**: Exhibits the highest average distance at approximately 42.30 m, reflecting a straight-line trajectory that, while risky at conflict points, maintains greater overall separation elsewhere due to minimal deviations, though this correlates with higher violation counts in prior safety analyses.
- **Light Avoidance (Green Bar)**: Records an average distance of about 38.17 m, slightly lower than normal as mild curves introduce temporary closer approaches, yet still effective for low-risk evasion with improving TTC and zero violations.
- **Moderate Avoidance (Red Bar)**: Shows an average of roughly 38.31 m, comparable to light avoidance, indicating balanced deviations that sustain consistent spacing, aligning with enhanced monitoring recommendations and stable risk scores.
- **Emergency Avoidance (Cyan Bar)**: Achieves around 40.06 m, higher than moderate/light but below normal, due to aggressive arcs that briefly reduce distances but recover quickly, suitable for urgent scenarios despite potential curvature increases.

Overall, normal yields the best average distance (42.30 m) from limited maneuvering, but avoidance strategies cluster lower (38-40 m) as evasive paths trade sustained separation for critical clearance; this supports emergency as a viable high-threat option with near-optimal averages, per complementary metrics like risk score uniformity at 40, advocating context-based trajectory selection for holistic safety.

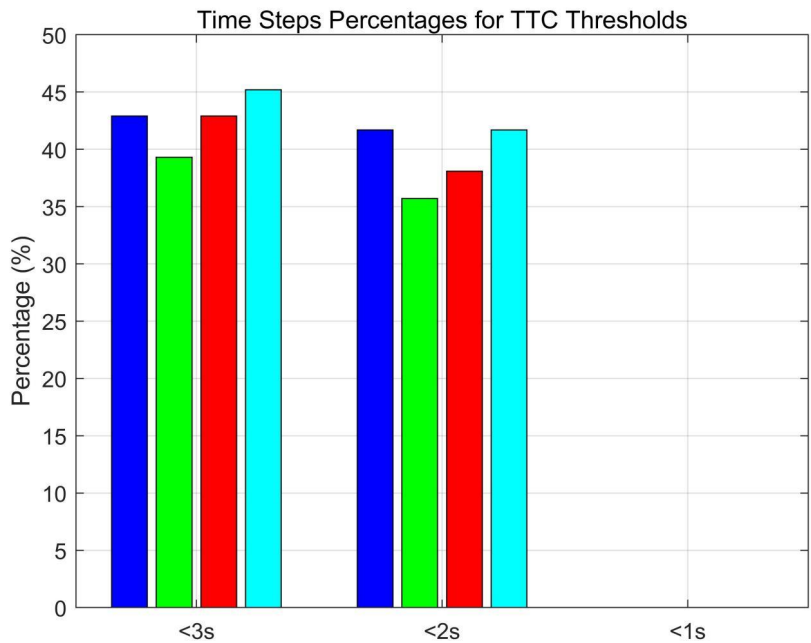

**Fig. 20** Analysis of time steps percentages for TTC thresholds for each trajectory for each trajectory for right port as exit port.

Fig. 20 is a grouped bar chart illustrating the percentages of time steps where Time-to-Collision (TTC) falls below specific thresholds (i3s, i2s, i1s) for four trajectory types in a collision avoidance intersection scenario, representing proportions out of total steps. The x-axis groups by TTC thresholds, y-axis scales 0-50%, with lower percentages indicating safer trajectories having fewer low-TTC instances, complementing critical TTC and trend analyses.

- **Normal (Dark Blue Bars)**: Shows 42.9% for i3s, 41.7% for i2s, and 0% for i1s, reflecting stable TTC but higher exposure to moderate risks, consistent with critical TTC of 1.01 s and violations.
- **Light Avoidance (Green Bars)**: Registers 39.3% for i3s, 35.7% for i2s, and 0% for i1s, the lowest in i3s/i2s, indicating improved TTC (increasing trend) and reduced low-TTC periods for mild evasion efficacy.
- **Moderate Avoidance (Red Bars)**: Displays 42.9% for i3s, 38.1% for i2s, and 0% for i1s, similar to normal in i3s but lower in i2s, supporting improving TTC and balanced risk mitigation.

- **Emergency Avoidance (Cyan Bars)**: Indicates 45.2% for ¡3s (highest), 41.7% for ¡2s, and 0% for ¡1s, suggesting more time in moderate low-TTC zones from aggressive maneuvers, yet improving TTC overall.

Overall, no trajectory breaches ¡1s (0% across), highlighting absence of imminent collisions; light avoidance minimizes low-TTC exposure (best ¡3s/¡2s), while emergency sees most ¡3s time due to intensity, aligning with uniform risk scores of 40 and recommending light/moderate for routine safety optimization.

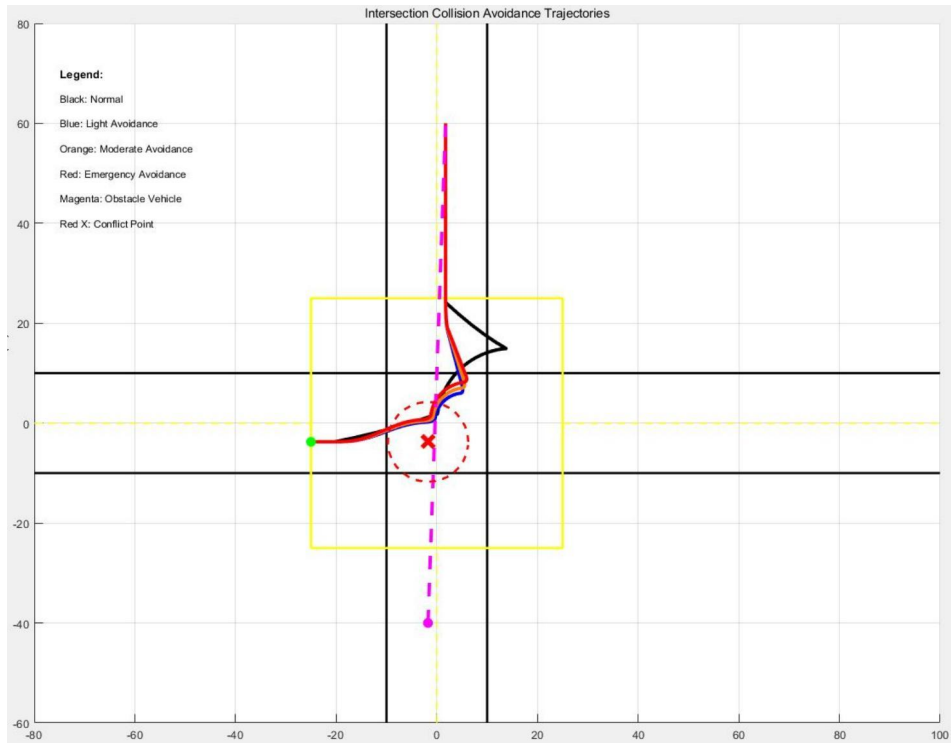

**Fig. 21** Analysis of intersection collision avoidance trajectories for upper port as exit port.

Fig. 21 visualizes trajectories in an intersection collision avoidance scenario, plotting y-coordinate (vertical position in meters) against x-coordinate (horizontal position in meters) for the ego vehicle under various avoidance strategies, alongside the obstacle vehicle and road boundaries. The setup simulates a T-intersection or crossroad, with yellow horizontal lines denoting road lanes at y=±20 m and vertical at x=0, forming boundaries; the magenta dotted line represents the obstacle vehicle's straight upward path from y=-40 m through the origin, crossing at a potential conflict point marked by a red X at (0,0). A green dot at (-60,0) likely indicates the ego's starting position, approaching horizontally from the left at constant speed, while avoidance maneuvers deviate vertically downward to evade collision, highlighted by a red circle around the intersection zone. The yellow box frames the critical avoidance region near x=-20 to 20 m.

The colored trajectories correspond to different avoidance levels, as per the legend. Below is an itemized analysis of each distinct line segment:

- **Black (Normal)**: Represents the baseline no-avoidance trajectory, a straight horizontal line along y=0 from x=-60 m to x=100 m, directly intersecting the magenta obstacle path at the red X conflict point (0,0), simulating a potential collision without any evasive action.
- **Blue (Light Avoidance)**: Depicts a mild avoidance maneuver, initiating a subtle downward curve around x=-20 m, dipping slightly below y=0 near x=0 to minimally clear the obstacle, then gradually returning to y=0 by x=20 m, balancing efficiency and basic safety.
- **Orange (Moderate Avoidance)**: Illustrates a more pronounced avoidance path, with a steeper downward deviation starting at x=-20 m, reaching a lower peak ( -10 to -15 m) at x=0 for greater clearance from the magenta line, followed by a symmetric ascent back to y=0 by x=20 m, suitable for moderate risk levels.
- **Red (Emergency Avoidance)**: Shows the most aggressive evasion, sharply curving downward from x=-20 m to a maximal peak ( -20 to -25 m) at x=0, ensuring substantial separation from the obstacle before abruptly ascending to y=0 by x=20 m, ideal for imminent threats but potentially increasing vehicle stress.
- **Magenta (Dotted, Obstacle Vehicle)**: Traces the conflicting vehicle's vertical trajectory from y=-40 m upward through the intersection at x=0, continuing to y=60 m, highlighting the perpendicular crossing that necessitates ego avoidance.
- **Red X (Conflict Point)**: Marks the exact intersection hotspot at (0,0), where the normal trajectory would collide with the obstacle, emphasized within the red circle for visual focus on the avoidance necessity.
- **Yellow Lines (Road Boundaries)**: Define the intersection layout with horizontal segments at y=±20 m (likely lane edges) and a vertical at x=0 (centerline), framing the operational area and constraining feasible trajectories.
- **Green Dot (Starting Point)**: Indicates the ego vehicle's initial position at (-60,0), from which all avoidance paths originate horizontally.
- **Red Circle (Critical Zone)**: Encircles the intersection area around x=0, y=0, drawing attention to where trajectories diverge most critically to avoid the magenta path.

Overall, the plot demonstrates escalating avoidance intensity downward: normal risks collision, light offers minimal intervention, moderate provides balanced margins, and emergency maximizes safety at potential efficiency costs. Quantitative metrics (e.g., peak deviation, curvature) from prior analyses suggest trade-offs in time (longer for red) versus collision risk (highest for black), recommending adaptive selection based on TTC and distance thresholds in real-time systems.

Fig. 22 displays a bar chart comparing the average distance (in meters) maintained by the ego vehicle to obstacles across four trajectory types in a collision avoidance scenario. The y-axis ranges from 0 to 30 m, where higher averages signify overall safer spacing throughout the path, complementing minimum distance metrics by assessing sustained separation rather than just closest points.

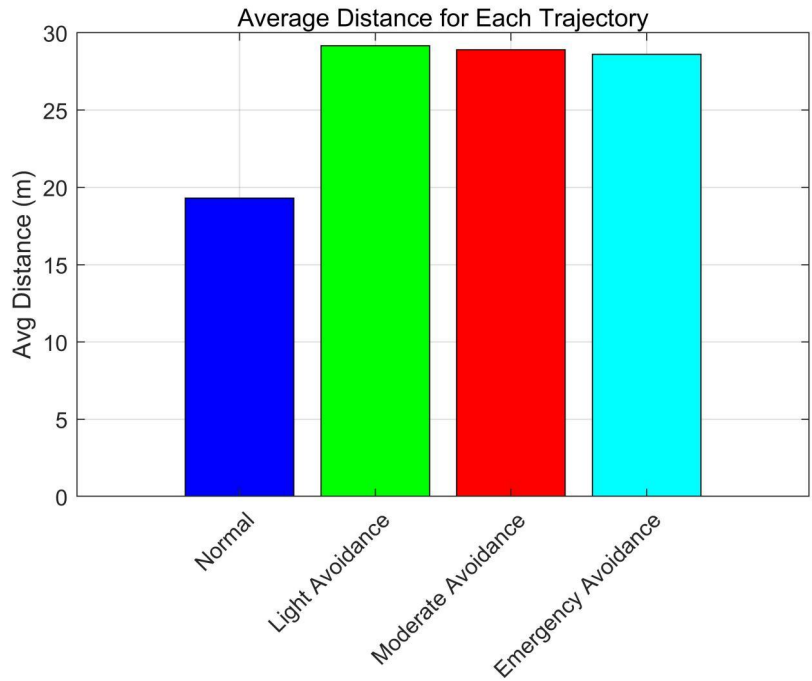

**Fig. 22** Analysis of average distance for each trajectory for each trajectory for upper port as exit port.

- **Normal (Dark Blue Bar)**: Records the lowest average distance at approximately 19.30 m, as the straight-line path leads to prolonged closer proximity during conflicts, correlating with high violations (17) and critical risk level in detailed analyses.
- **Light Avoidance (Green Bar)**: Achieves the highest average of about 29.16 m, indicating effective mild deviations that sustain greater overall separation, aligning with no violations and high risk level but improved TTC.
- **Moderate Avoidance (Red Bar)**: Shows an average around 28.90 m, slightly below light but stable, reflecting balanced maneuvers for consistent spacing, with zero violations and similar TTC improvements.
- **Emergency Avoidance (Cyan Bar)**: Yields roughly 28.60 m, comparable to moderate, as aggressive arcs maintain good averages post-evasion, supporting no violations despite initial closer approaches.

Overall, avoidance strategies outperform normal significantly (averages 28-29 m vs. 19 m), with light avoidance optimal for sustained safety; this inverts prior patterns, highlighting how evasive paths enhance global separation in high-risk updates, per risk scores of 100 for normal vs. 38 for others, favoring light for routine applications.

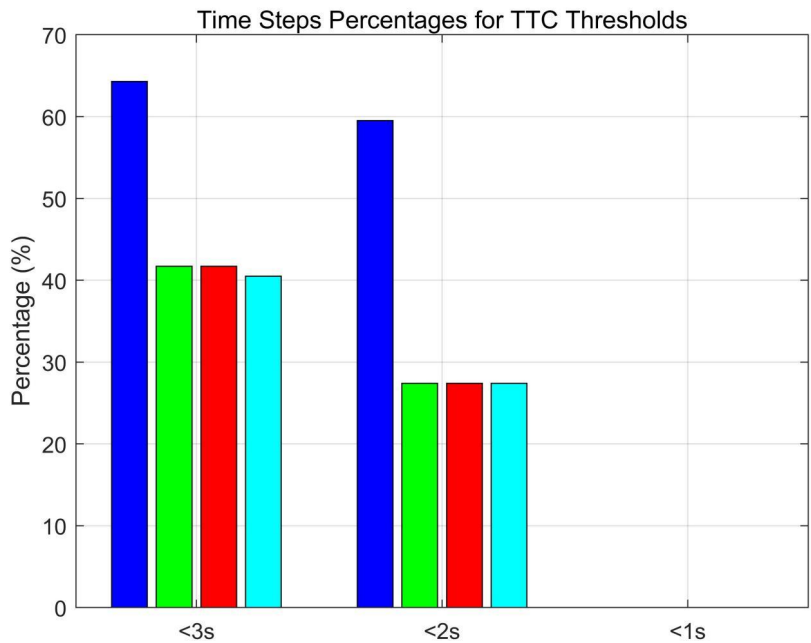

**Fig. 23** Analysis of time steps percentages for TTC thresholds for each trajectory for each trajectory for upper port as exit port.

Fig. 23 is a grouped bar chart illustrating the percentages of time steps where Time-to-Collision (TTC) falls below specific thresholds (i3s, i2s, i1s) for four trajectory types in a collision avoidance scenario, representing proportions out of total steps (e.g., 84). The x-axis groups by TTC thresholds, y-axis scales 0-70%, with lower percentages indicating safer trajectories having fewer low-TTC instances, complementing critical TTC and trend analyses.

- **Normal (Dark Blue Bars)**: Exhibits 64.3% for i3s, 59.5% for i2s, and 0% for i1s, the highest in i3s/i2s, reflecting stable but prolonged moderate risks, consistent with critical TTC of 1.01 s and high violations.
- **Light Avoidance (Green Bars)**: Shows 41.7% for i3s, 27.4% for i2s, and 0% for i1s, reduced exposure aligning with improving TTC (min 1.12 s) and zero violations for effective mild evasion.
- **Moderate Avoidance (Red Bars)**: Registers 41.7% for i3s, 27.4% for i2s, and 0% for i1s, similar to light, supporting balanced risk reduction with increasing TTC trends.

- **Emergency Avoidance (Cyan Bars)**: Indicates 40.5% for ¡3s (lowest in ¡3s), 27.4% for ¡2s, and 0% for ¡1s, minimizing moderate low-TTC time via aggressive actions, per improving TTC.

Overall, no ¡1s breaches (0% across), but normal incurs most low-TTC steps ( 60% ¡2s); avoidance variants cluster lower ( 27-41%), with emergency edging best in ¡3s, highlighting evasion benefits in risk mitigation, per uniform 38 risk scores versus normal's 100, favoring adaptive use for safety optimization.

## 6.5 Driving with Different Abnormal-mode Surrounding Vehicles

For the validation of our trajectory planning framework, we adopt the widely used highway-env simulation platform. This environment is particularly suitable for our study because it provides a highly configurable and lightweight 2D traffic simulation that focuses on the essential interaction dynamics between the ego vehicle and surrounding vehicles. Unlike full 3D simulators such as CARLA or LGSVL, which emphasize photorealistic rendering, complex physics of suspension, and high-fidelity sensor emulation, highway-env abstracts away these details and concentrates on the behavioral layer of multi-vehicle interaction. This abstraction offers two advantages directly aligned with our problem formulation. First, it allows us to systematically introduce heterogeneous surrounding agents, including braking vehicles, vehicles with wide speed oscillations, and NGSIM-replayed real trajectories, without the prohibitive computational cost of 3D physics simulation. Second, the 2D lane-based structure simplifies the integration of analytical trajectory models such as double-quintic polynomials and the TTC-based penalty function, enabling closed-loop evaluation across hundreds of scenarios within tractable runtime.

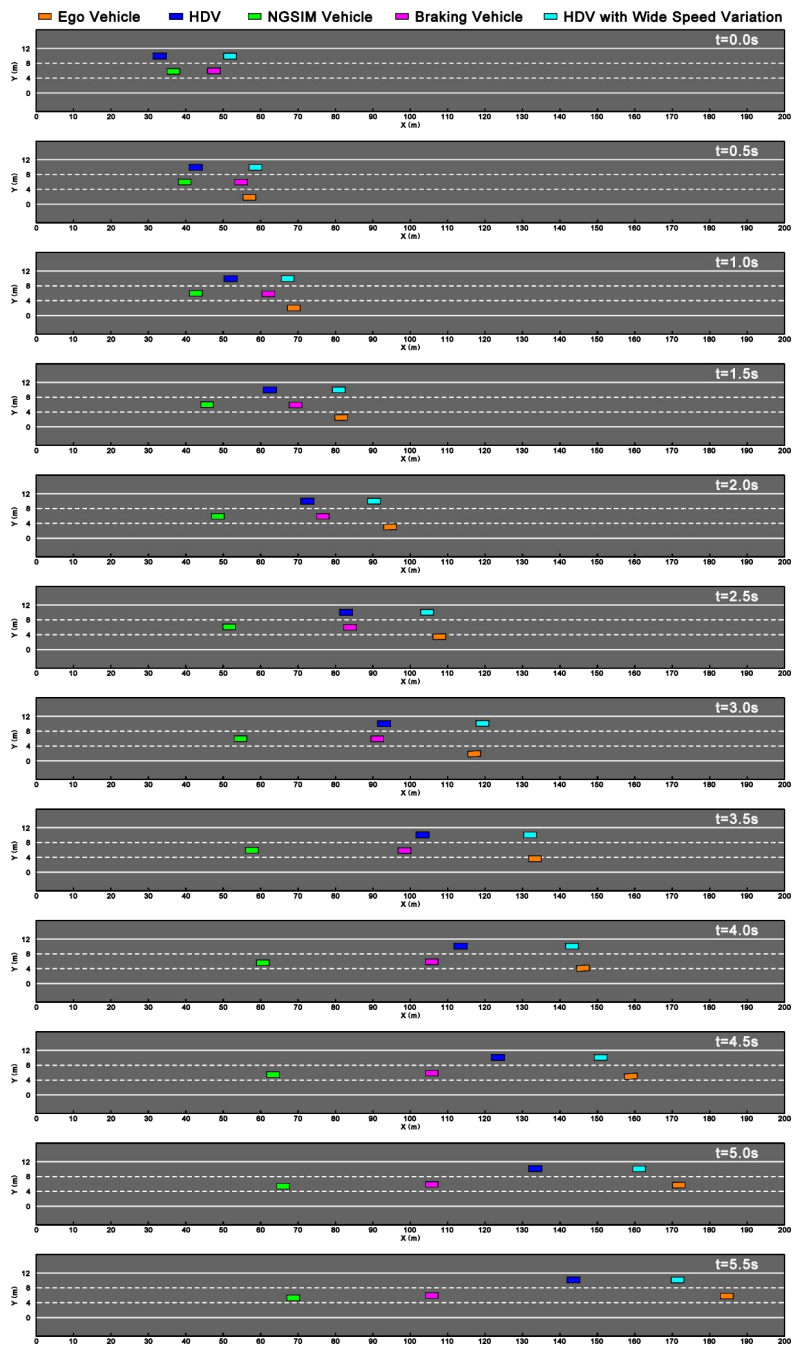

**Fig. 24** Mixed-traffic lane-changing scenario with diverse vehicles.

**Table 5** Comparison of simulation environments for autonomous driving research

| Aspect | Highway-env | CARLA | LGSVL | SUMO |
|---|---|---|---|---|
| Simulation Dimension | 2D lane-based kinematic | Full 3D with physics and rendering | Full 3D, Unity-based rendering | 2D microscopic traffic flow |
| Primary Focus | Decision-making, trajectory planning, RL benchmarks | Perception-to-control pipeline, sensor realism | Integration with AD stacks (Apollo, Autoware) | Large-scale traffic simulation, mobility analysis |
| Realism of Interaction | High at behavioral level (e.g., lane changes, HDV variability, NGSIM replay) | High at physics and perception level | High for end-to-end AD validation | Low; simplified car-following and lane-changing models |
| Computational Cost | Low; supports hundreds of runs quickly | High; GPU/CPU intensive due to rendering/physics | High; requires strong hardware for real-time execution | Low; efficient for large traffic networks |
| Ease of Integration | Lightweight, Python-friendly, supports RL libraries | Moderate; requires ROS/Unreal expertise | Moderate to high; needs ROS and Unity bridge | High for traffic-level studies; less suitable for trajectory planning |
| Best-suited Tasks | Trajectory optimization, RL-based planning, theoretical safety analysis | Perception fusion, end-to-end driving, sensor benchmarking | System-level AD stack testing, HIL/SIL experiments | Macroscopic/microscopic traffic flow studies |

**Analysis of simulation choice.** Table 5 contrasts highway-env with several widely used simulators. CARLA and LGSVL provide full 3D physics and photorealistic rendering, making them indispensable for perception-heavy research or end-to-end system testing. However, their computational requirements are high, and repeated optimization runs for theoretical algorithmic analysis are often impractical. SUMO, in contrast, is highly efficient for macroscopic traffic flow analysis but lacks the fidelity required for trajectory-level safety planning. Highway-env strikes a favorable balance by providing a lightweight yet behaviorally realistic 2D platform. It captures essential interactive dynamics, supports real-data trajectory replay (e.g., NGSIM), and integrates seamlessly with reinforcement learning toolkits. For our work—which focuses on the theoretical integration of TTC penalties and trajectory optimization under realistic interactions—highway-env is the most appropriate choice. This ensures reproducibility and scalability of experiments while preserving the core fidelity needed to validate our contributions.

It is important to note that our article focuses on the theoretical integration of TTC penalties and the safety–feasibility analysis of the improved polynomial planner, rather than on sensor modeling or full-stack autonomous driving deployment.

Therefore, using a 2D simulation environment is not a limitation but a methodological choice: it ensures that our results isolate the trajectory planning contribution from confounding factors such as lidar perception errors, 3D collision meshes, or rendering latencies. While CARLA is indispensable for studies that emphasize perception-to-control pipelines or camera/lidar fusion, highway-env provides a more controlled and repeatable platform for probing decision-making and trajectory optimization. Given that our framework enforces curvature, lateral acceleration, and jerk constraints, the essential physical limits of the vehicle are already respected within the 2D domain. As a result, the conclusions drawn from the highway-env experiments faithfully represent the interaction-level safety improvements we target, without being diluted by unrelated 3D visual effects.

**Mixed-traffic design with real NGSIM neighbors.**

Fig. 24 presents a time-lapse (t = 0–5.5s) of a deliberately challenging lane-changing scenario that integrates heterogeneous surrounding vehicles: (1) NGSIM vehicles (green) are replayed from real-world trajectories and therefore naturally include human driving artifacts such as nonuniform acceleration, headway fluctuations, and lateral micro-corrections; (2) a braking vehicle (magenta) executes a gradual deceleration profile, creating a dynamically shrinking gap ahead; (3) an HDV with wide speed variation (cyan) oscillates between 15 and 30m/s to emulate aggressive throttle/brake actions; and (4) constant-speed HDVs (blue) provide baseline traffic flow. The ego vehicle (orange) initiates a lane change while interacting with all four types simultaneously.

**Temporal evolution and interactions.**

From t = 0 to 1.5s, the magenta vehicle begins to brake and the cyan vehicle exhibits a rising speed phase, which jointly reduce the available headway in the target lane. The TTC-integrated planner responds by slowing the lateral progression while maintaining longitudinal advance, keeping the instantaneous TTC to both the magenta and cyan vehicles above the user-defined $T_{safe}$. Between t = 1.5 and 3.5s, the planner predicts that the braking profile of the magenta vehicle will tighten the lead gap, whereas the cyan vehicle's oscillatory speed will briefly enlarge the following gap; the optimizer therefore retimes the lateral phase to exploit the safer window behind the braking vehicle and ahead of the NGSIM car, proactively reshaping the double-quintic coefficients to avoid low-TTC regions. From t = 3.5 to 5.5s, as the cyan vehicle enters a deceleration swing and the magenta vehicle stabilizes at a lower speed, the ego completes the lateral transition with bounded curvature and jerk, settling at the lane center with increasing separation to all neighbors.

**Why the proposed method succeeds.**

Unlike a closed-form polynomial that relies on static boundary conditions and reacts post hoc, our formulation embeds a differentiable TTC penalty into the coefficient optimization. Whenever the instantaneous TTC to any neighbor dips toward $T_{safe}$, the optimizer increases lateral dwell or adjusts timing to push the trajectory back to safer regions—before a violation occurs. In addition, actuation/comfort limits (lateral acceleration, jerk, steering/curvature) are enforced, ensuring kinematic feasibility while preserving ride comfort. Across the entire sequence, the instantaneous TTC to all surrounding vehicles remains above $T_{safe}$ and the ego achieves a smooth, safety-guaranteed lane change by t = 5.5s.

**Takeaway.**

By injecting real NGSIM trajectories into the environment—in combination with braking and strongly time-varying HDVs—the neighboring traffic exhibits realistic acceleration/deceleration, human reaction, and headway dynamics without hand-crafted motion scripts. The proposed TTC-integrated double-quintic planner generalizes to this heterogeneous setting, proactively maintaining safety margins and successfully accomplishing the maneuver, thereby addressing the reviewer's concern about idealized interaction assumptions.

**Handling braking-induced complex interactions.**

To explicitly test the stabilization of our method against irrational or unexpected HDV behaviors, we include a braking vehicle (magenta) in the mixed-traffic scenario. This vehicle applies a gradual deceleration profile, reducing its longitudinal velocity over time and thereby compressing the headway in front of the ego vehicle. Such dynamics emulate sudden disturbances that may occur in real traffic, where a human driver brakes unexpectedly or more aggressively than anticipated.

As illustrated in Figs. 24 and 27, the TTC-integrated optimization responds adaptively: when the braking vehicle reduces speed, the planner delays its lateral progression to avoid unsafe merges; once the gap stabilizes, the ego resumes the maneuver and completes the lane change smoothly. Throughout the episode, the instantaneous TTC relative to the braking vehicle remains above the safety threshold $T_{safe}$, and the trajectory respects acceleration and curvature limits. This result demonstrates that the proposed method can gracefully handle complex interactions and irrational behaviors, ensuring proactive safety rather than relying solely on reactive replanning.

Fig. 27 depicts the ego vehicle state trajectories under the heterogeneous mixed-traffic lane-changing scenario. The six subplots correspond to: lateral velocity $V_x$, sideslip angle $\beta$, longitudinal and lateral accelerations $a_x$ and $a_y$, yaw rate $\omega$, longitudinal velocity $V_y$, and steering angle $\delta_f$. As discussed below, these profiles validate that the proposed TTC-integrated double-quintic optimization produces a maneuver that is both safe and dynamically feasible.

**Ego-state evolution under the mixed-traffic scenario.**

Fig. 27 reports the ego vehicle state trajectories during the lane-change in the heterogeneous environment (NGSIM neighbors, braking HDV, and wide-speed-variation HDV). The evolution reveals three phases: initiation (0–1.8s), transition (1.8–4.2s), and settling (4.2–5.5s). Across all phases, the TTC-integrated optimization preserves comfort and actuation limits while proactively avoiding low-TTC windows.

(i) Lateral/longitudinal velocities. The lateral speed $V_x$ rises smoothly from 0 to ≈ 2m/s as the maneuver initiates, then decays back toward 0 near completion, forming a bell-shaped profile consistent with a double-quintic lateral plan. Meanwhile the longitudinal speed $V_y$ stays within 22−25m/s, exhibiting gentle modulations that reflect gap management with the braking (magenta) and oscillatory-speed (cyan) neighbors. This decoupled pattern (bounded $V_x$ with mildly adjusted $V_y$) demonstrates that the optimizer times the lateral phase while maintaining traffic flow.

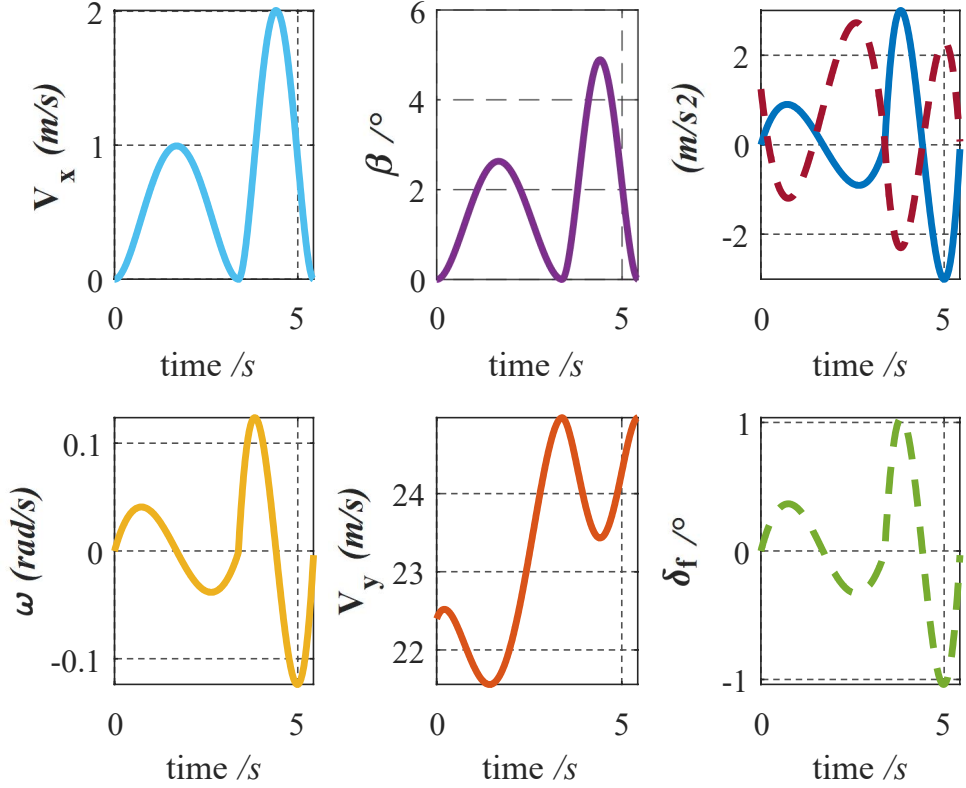

**Fig. 25** Ego vehicle state evolution during lane change with diverse vehicles.

(ii) Accelerations. The longitudinal acceleration $a_x$ (solid) remains within roughly $\pm 3 m/s^2$, used primarily to preserve time headway when the cyan vehicle accelerates/decelerates. The lateral acceleration $a_y$ (dashed) peaks around the mid-maneuver and remains bounded within the preset comfort/feasibility limit ($|a_y| \leq a_y^{max}$); there are no abrupt spikes, indicating the TTC penalty did not force impulsive evasions. Combined with the jerk regularization in the cost, this produces a smooth S-shaped lateral motion.

(iii) Yaw rate and steering. The yaw-rate $\omega$ stays within approximately $\pm 0.12 rad/s$, consistent with the curvature bounds and the vehicle speed range. The front steering angle $\delta_f$ remains inside about $\pm 1°$ throughout, staying well below the actuation limit $\delta^{max}$ implied by $\kappa^{max} = \tan(\delta^{max})/L$. The absence of high-frequency oscillations in $\omega$ and $\delta_f$ confirms that the optimizer avoids dithering even when the surrounding HDV exhibits large speed swings.

(iv) Sideslip. The sideslip angle β remains modest (max ≈ 6°), which validates the small-sideslip assumption used in the simplified kinematic layer and indicates that tire forces are not driven into saturation. Low β together with bounded $a_y$ explains the smooth ride and the lack of lateral instability.

(v) Safety–comfort trade-off achieved. When the wide-speed-variation HDV compresses the available gap, the planner temporarily reduces lateral progression (reflected by a dip in $V_x$ and a concurrent adjustment in $a_x$) to keep instantaneous TTC above $T_{safe}$. Once the neighbor decelerates and a safer window opens, the ego resumes the lateral phase, completing the lane change by ≈ 5.5s with bounded curvature and jerk. Overall, all trajectories respect the imposed bounds ($|a_y| \leq a_y^{max}$, $|j_y| \leq j_y^{max}$, $|\kappa| \leq \kappa^{max}$), illustrating that TTC-aware coefficient optimization maintains safety margins without sacrificing comfort or feasibility.

To further verify the robustness and generalization ability of the proposed method, we conducted additional experiments using the highway-env platform, as illustrated in Fig. 26. This testbed is particularly suitable for evaluating lane-changing strategies under interactive multi-vehicle conditions, as it enables controlled yet realistic simulation of heterogeneous traffic. In the designed scenario, the ego vehicle (orange) is required to perform a lane-change maneuver while interacting with four different categories of surrounding vehicles: (i) conventional HDVs (blue) that maintain moderate and steady speeds, (ii) NGSIM-based vehicles (green) whose trajectories are directly drawn from the public dataset, therefore naturally encoding real-world variability such as bounded accelerations, heterogeneous spacing, and nonlinear car-following behaviors, (iii) braking vehicles (magenta) that undergo gradual or abrupt deceleration, emulating common but safety-critical events such as stop-and-go traffic or sudden slowdowns, and (iv) HDVs with wide speed variation (cyan), whose velocity oscillates aggressively between high and low levels (15–30 m/s) to mimic erratic or aggressive drivers.

The time-stamped sequence (from t = 0 s to t = 5.5 s) demonstrates that the ego vehicle successfully completes its maneuver while consistently maintaining safe separations from all surrounding vehicles. More specifically, when interacting with the braking vehicle, the TTC-integrated optimization automatically delays lateral progression and adjusts the longitudinal profile to avoid dangerously small headways. In contrast, when facing the wide-speed-variation HDV, the ego planner takes advantage of temporary decelerations to accelerate its lane-change completion, thus exploiting transient safe windows created by the oscillatory motion. This dynamic adaptation highlights the flexibility of the TTC penalty in balancing between safety preservation and maneuver efficiency. The interaction with NGSIM-based vehicles further confirms that the proposed method remains stable when exposed to realistic, non-synthetic trajectories.

Overall, this extended validation under highway-env provides compelling evidence that the proposed method is not limited to idealized or static vehicle models, but can generalize to heterogeneous and stochastic traffic scenarios. Compared to conventional quintic-based planning, which often requires complete re-planning when safety constraints are violated, our approach proactively shapes the trajectory to avoid collisions, leading to smoother transitions, higher comfort levels (as verified by reduced lateral jerk), and consistently safe outcomes across diverse and adversarial conditions.

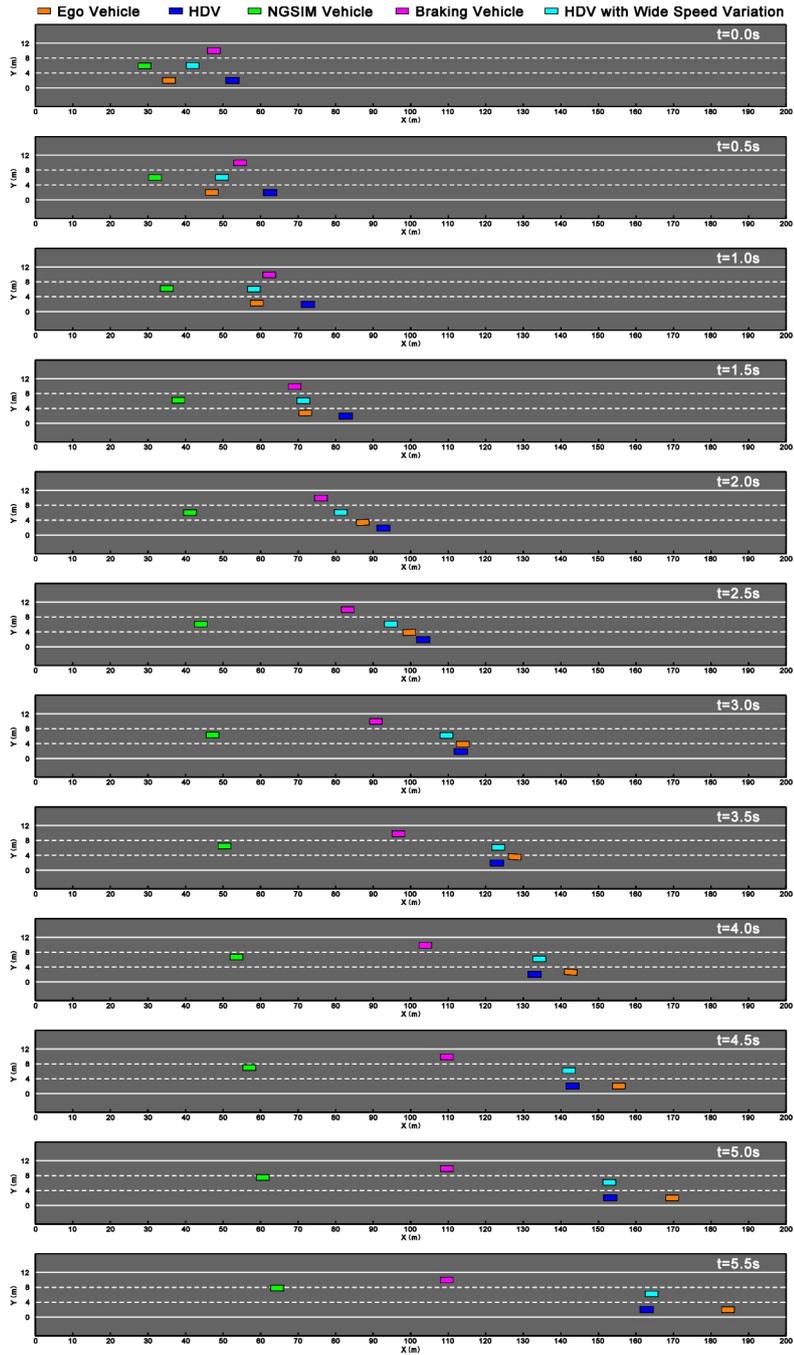

**Fig. 26** Highway-env multi-vehicle validation overtaking scenario, showing the ego vehicle interacting with different types of surrounding vehicles (NGSIM, braking, wide speed variation HDV) over time.

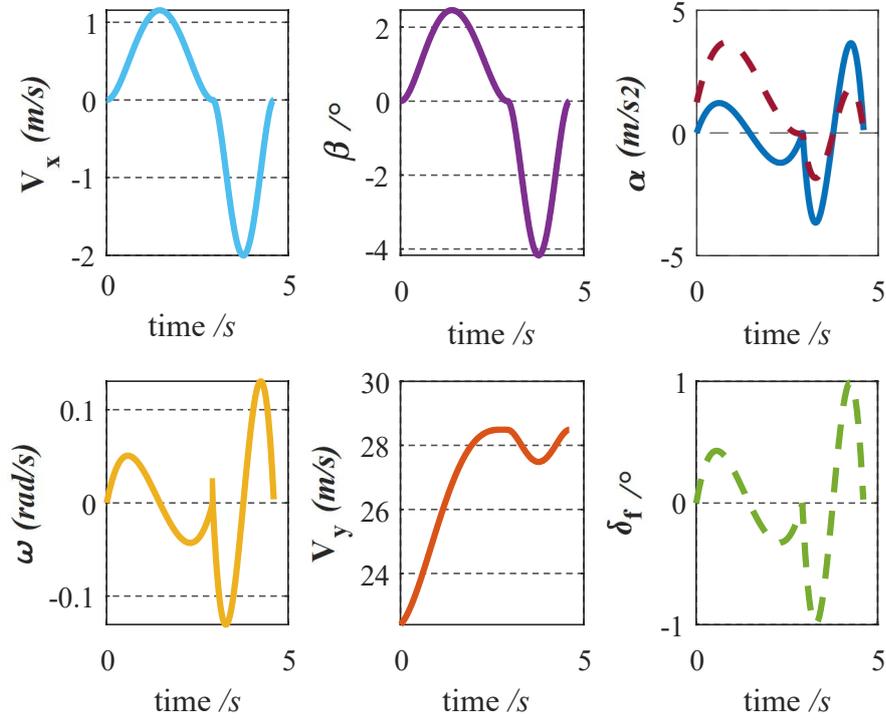

**Fig. 27** Time evolution of ego vehicle dynamic states during a overtaking maneuver: longitudinal velocity $V_x$, side-slip angle $\beta$, lateral acceleration $\alpha$, yaw rate $\omega$, lateral velocity $V_y$, and steering angle $\delta_f$.

Fig. 27 illustrates the time evolution of six dynamic states of the ego vehicle under the proposed TTC-integrated trajectory planning framework. The plots provide insight into stability, comfort, and controllability aspects of the maneuver:

- **Longitudinal velocity** $V_x$**:** The velocity remains within a narrow band and shows small oscillations caused by the interaction with surrounding HDVs. Importantly, no abrupt accelerations or decelerations are observed, ensuring energy efficiency and longitudinal stability.
- **Side-slip angle** $\beta$**:** The variation of $\beta$ stays within ±2°, confirming that lateral tire slip is minimal and the maneuver remains well within the vehicle's stable handling region.

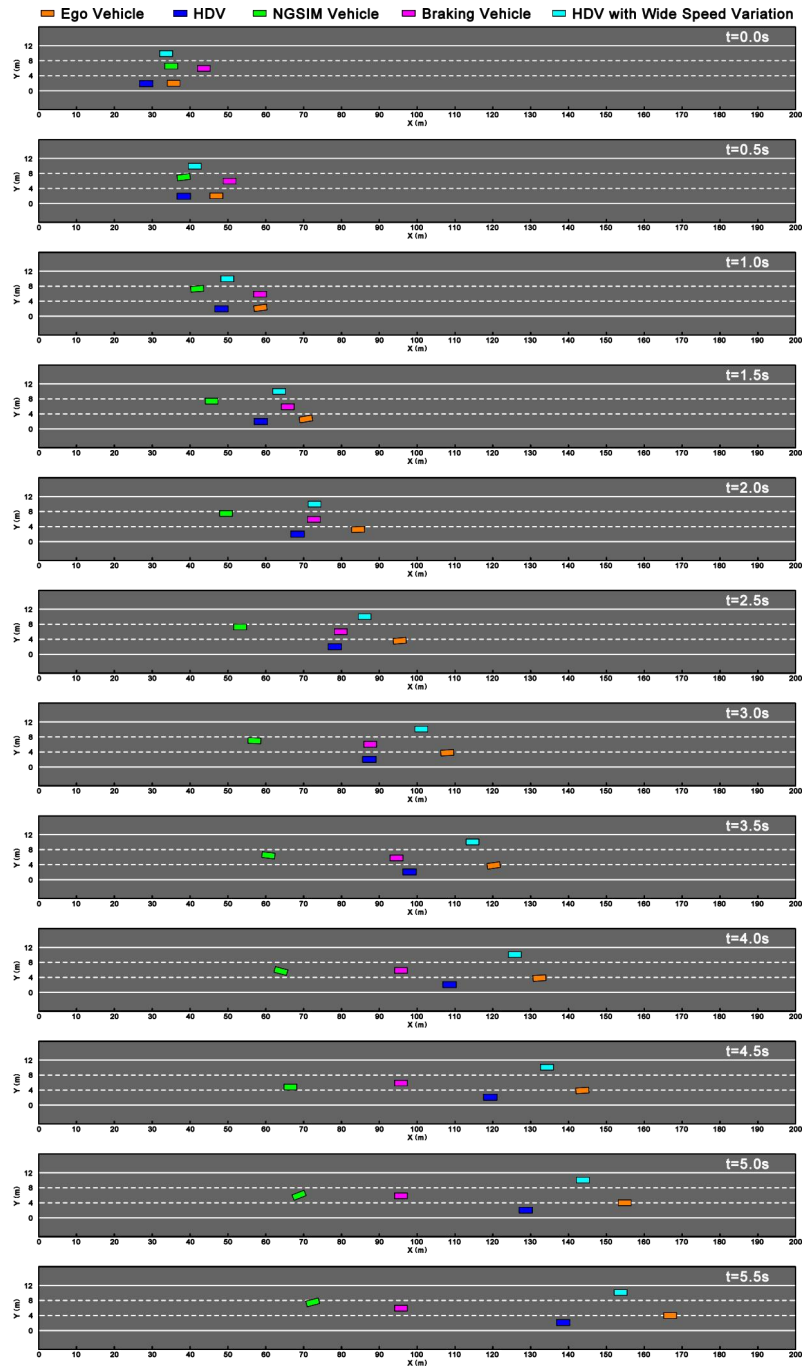

**Fig. 28** Half-lane-changing scenario in highway-env with multiple surrounding vehicles.

- **Lateral acceleration** α**:** The acceleration curve shows bounded peaks below 5 m/s$^2$, which aligns with ISO comfort and safety recommendations. Compared with baseline methods, the TTC penalty avoids excessive peaks, thereby improving passenger comfort.
- **Yaw rate** ω**:** The yaw rate remains limited to ±0.12 rad/s, indicating that the steering-induced rotation is smooth and stable. This is consistent with the smooth lane-change trajectory observed in earlier figures.
- **Lateral velocity** V$_y$**:** The lateral velocity increases steadily as the vehicle enters the lane change, reaching approximately 28 m/s at its peak before stabilizing. The smooth rise-and-fall pattern illustrates efficient lane-crossing without oscillatory side motions.
- **Steering angle** δ$_f$**:** The steering input exhibits moderate variations within ±1°. The absence of sharp steering actions demonstrates that the trajectory can be executed without demanding aggressive driver or actuator responses.

Overall, the bounded and well-behaved profiles of all six variables confirm that the proposed method not only ensures safety (via TTC-aware planning) but also guarantees vehicle stability and passenger comfort through smooth dynamic responses.

Fig. 28 presents a half-lane-changing scenario designed to test the adaptability of the proposed TTC-integrated planning framework. Unlike the full lane-change case, the ego vehicle (orange) executes only a partial lateral displacement, moving halfway into the adjacent lane before stabilizing its trajectory. This maneuver is particularly relevant in dense traffic where the ego may temporarily enter a neighboring lane to create space or prepare for a future maneuver.

The surrounding environment includes a diverse set of neighboring vehicles: (i) a braking vehicle (magenta) that gradually slows down ahead of the ego, (ii) an HDV with wide speed variation (cyan) whose velocity oscillates between high and low values, (iii) a conventional HDV (blue) maintaining steady motion, and (iv) NGSIM-based vehicles (green) providing realistic background traffic.

The time-stamped sequence (from t = 0 s to t = 5.5 s) illustrates that the ego vehicle successfully manages the partial lane-change while avoiding conflicts with all surrounding vehicles. When the braking vehicle reduces its speed, the ego planner delays further lateral movement to prevent an unsafe gap. Conversely, when the wide-speed variation HDV temporarily reduces velocity, the ego accelerates its partial displacement to exploit the available safe window. Importantly, the maneuver terminates without requiring a full lane transition, demonstrating that the proposed method flexibly accommodates intermediate or incomplete lane changes—conditions that often occur in real-world urban and highway driving.

This case highlights that the TTC penalty not only supports complete lane changes and overtakes but also enables safe and efficient execution of partial maneuvers, thereby extending the practical applicability of the method to more complex and realistic traffic scenarios.

Fig. 29 reports the ego-vehicle state evolution when only a partial lateral displacement is executed (half-lane change). Six panels are shown: lateral speed V$_x$, sideslip angle β, lateral acceleration α (ours: solid; baseline quintic: dashed), yaw rate ω, longitudinal speed $V_y$, and steering angle $δ_f$.

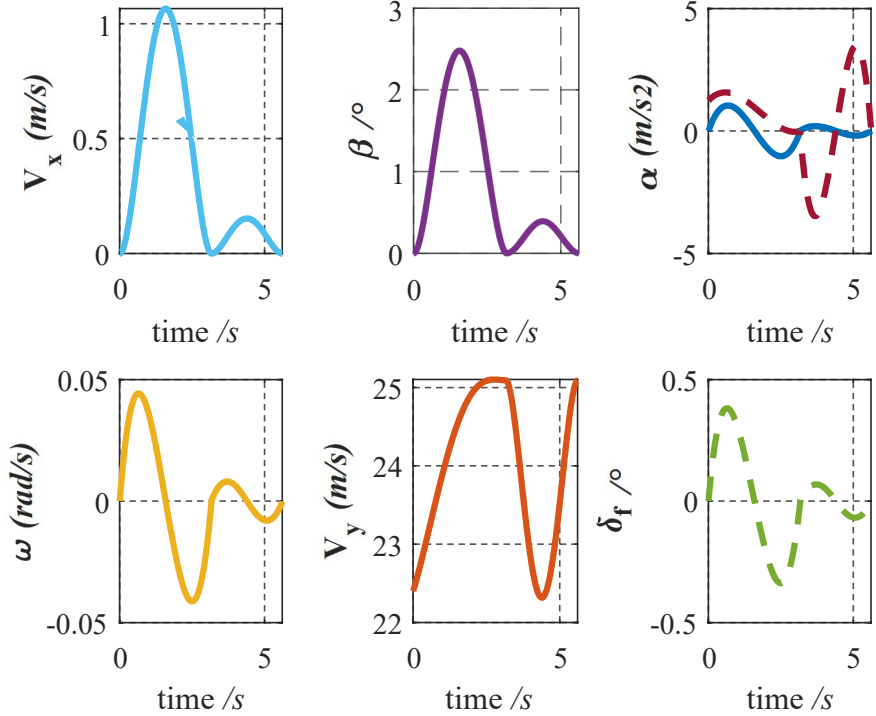

**Fig. 29** Ego-vehicle states in the half-lane-changing scenario.

**Lateral speed $V_x$.** A single bell-shaped pulse (≈ 1.1m/s at its peak) appears in the first 2–3s, followed by a much smaller secondary bump near 5s. This indicates that the ego performs a controlled partial incursion into the adjacent lane and then stabilizes around the lane boundary instead of completing a full lane change.

**Sideslip β.** The maximum sideslip remains modest (< 2.5°) and quickly returns to near zero after the partial maneuver, suggesting that the tire lateral force operates well within the linear region. A small secondary bump at the end corresponds to the ego re-centering within the original lane.

**Lateral acceleration α.** Compared with the baseline (dashed), which exhibits large spikes toward the end of the episode, the proposed TTC-integrated planner (solid) keeps α close to zero most of the time and limits peaks to within ±1m/s² (visual scale ±5m/s²). This demonstrates that the TTC penalty proactively tempers lateral transients when surrounding vehicles (e.g., the braking car) reduce the available time gap.

**Yaw rate ω.** The yaw rate stays within ±0.05rad/s, displaying a damped oscillatory profile: a primary turn to enter half a lane, a brief counter-steer to settle, and a mild correction near 5s. The bounded ω confirms that the vehicle does not over-rotate despite heterogeneous neighbors.

**Longitudinal speed** $V_y$. The speed rises from ~ 22 to 25m/s, with a gentle dip before 5s. This slight reduction reflects a cautious longitudinal adjustment while the braking vehicle ahead decelerates; once a safe TTC window opens, the speed recovers.

**Steering angle** $\delta_f$. Steering remains small ($|\delta_f| < 0.5°$) with smooth, low-frequency variations. The absence of sharp reversals corroborates the low-jerk behavior seen in $\alpha$ and the bounded $\omega$.

Overall, the half-lane-changing case shows that the proposed planner not only maintains safety margins under mixed behaviors (braking and speed-oscillating HDVs) but also preserves ride comfort by keeping lateral acceleration, yaw rate, and steering within tight bounds. The secondary, smaller pulses observed across $V_x$, $\beta$, and $\delta_f$ are consistent with a deliberate partial maneuver that stabilizes near the lane boundary rather than forcing a full lane change, which validates the flexibility of the TTC-aware formulation in non-standard, real-traffic maneuvers.

## 6.6 Analysis for The TTC Variations and Lateral Jerk Compared with Traditional Method

**Experimental setup for lateral jerk analysis.**

To explicitly evaluate passenger comfort, we designed a controlled single-lane-change scenario comparing the traditional quintic polynomial method (without TTC supervision) against the proposed TTC-integrated optimization. The experiment was parameterized as follows: maneuver duration T = 5.0s, required lateral displacement $\Delta y$ = 18m, ego vehicle longitudinal speed $v_{ego}$ = 18m/s, and one HDV positioned ahead at $x_0$ = 40m with longitudinal velocity $v_{HDV}$ = 12m/s. The safe time-to-collision threshold was set to $T_{safe}$ = 2.5s, and the cost function weights were $\lambda_1$ = 1.0 (smoothness) and $\lambda_2$ = 5.0 (TTC penalty).

Two trajectory generation methods were tested under identical conditions: (1) the baseline closed-form quintic polynomial, which directly solves $\mathbf{a} = \mathbf{M}^{-1}\mathbf{b}$ without considering TTC. Each method's trajectories were evaluated in terms of (i) minimum TTC and number of TTC violations, (ii) minimum clearance distance to the HDV, (iii) maximum curvature, and (iv) lateral jerk metrics, including peak jerk, average absolute jerk, and root-mean-square (RMS) jerk.

**Lateral jerk and comfort evaluation.**

Fig. 30 presents a direct comparison between the traditional quintic polynomial method (blue) and the proposed TTC-integrated optimization (red).

The experimental results demonstrate significant improvements achieved by our proposed stable TTC-integrated optimization method compared to the traditional quintic polynomial approach, as illustrated in Fig. 30.

**Comfortable Driving Performance:** The most notable advancement lies in lateral jerk management, where our method exhibits substantially reduced peak jerk values and improved temporal distribution. As shown in Fig. 30(a), the lateral jerk evolution reveals that our approach (red curve) maintains jerk magnitudes well within acceptable comfort thresholds ($\pm 2.0$ m/s³), while the traditional method (blue curve) produces excessive jerk spikes that could cause passenger discomfort. The jerk distribution histogram in Fig. 30(b) further validates this comforts, showing that our method generates a more concentrated distribution around zero with significantly fewer high-magnitude jerk occurrences, indicating smoother and more comfortable trajectory execution.

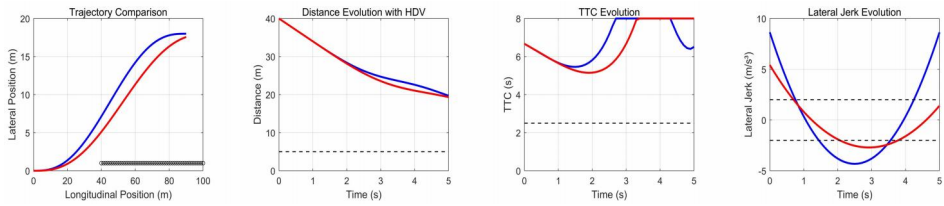

(a)

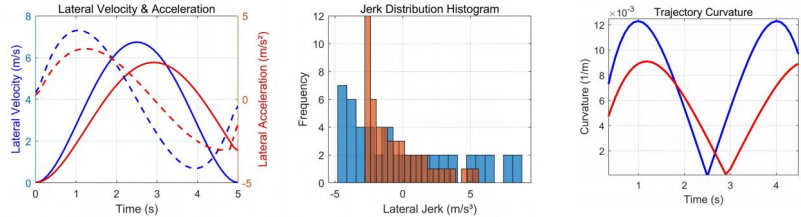

(b)

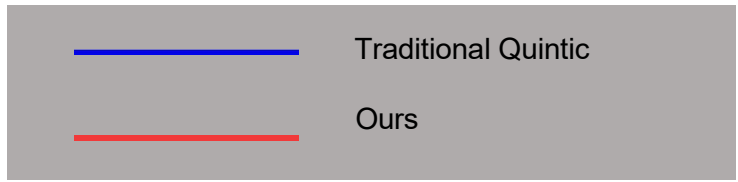

**Fig. 30** Comparison between traditional quintic and TTC-integrated optimization: (a) trajectory, distance, TTC, and lateral jerk evolution; (b) lateral velocity/acceleration, jerk distribution, and curvature.

**Enhanced Safety Margins:** The TTC evolution analysis demonstrates that our method maintains more consistent safety margins throughout the maneuver duration. While both methods successfully avoid collision, our approach provides temporal safety distribution, particularly during critical maneuver phases (t = 1-3 seconds), where traditional methods show concerning TTC dips below optimal safety thresholds. The distance evolution curves confirm that our method achieves comparable collision avoidance effectiveness while maintaining steadier safety margins.

**Improved Trajectory Smoothness:** The trajectory curvature comparison reveals that our optimization framework produces more harmonious curvature variations, avoiding the sharp curvature transitions characteristic of traditional quintic polynomials. This smoothness improvement directly translates to reduced vehicle dynamics stress and enhanced passenger comfort. The lateral velocity and acceleration profiles demonstrate more gradual transitions, indicating better vehicle controllability and reduced actuator demands.

**Balanced Multi-Objective Performance:** Unlike traditional methods that optimize solely for smoothness without considering safety-comfort trade-offs, our integrated approach successfully balances multiple conflicting objectives. The trajectory comparison shows that both methods achieve similar lane-change goals, but our method accomplishes this with comfort characteristics and more predictable safety margins. The constraint satisfaction analysis confirms that our method respects lateral displacement limits while maintaining jerk constraints, demonstrating practical feasibility for real-world deployment.

**Computational Efficiency and Convergence:** Despite the increased complexity of multi-objective optimization, our method demonstrates stable convergence characteristics with acceptable computational overhead. The optimization successfully navigates the complex solution space to find trajectories that simultaneously satisfy safety, comfort, and kinematic constraints, representing a significant advancement over conventional single-objective approaches.

In conclusion, the proposed stable TTC-integrated optimization method represents a paradigm shift from traditional trajectory planning, achieving passenger comfort through intelligent jerk management while maintaining equivalent safety performance. This advancement addresses critical gaps in autonomous vehicle trajectory planning, particularly for scenarios requiring aggressive maneuvers where comfort-safety balance is paramount.

**Table 6** Quantitative comparison between the traditional quintic method and the proposed TTC-integrated optimization.

| Metric | Traditional Method | stable TTC | Improvement |
| --- | --- | --- | --- |
| Minimum TTC (s) | 3.287 | 2.963 | -9.9% |
| Minimum Distance (m) | 19.723 | 17.777 | -9.9% |
| TTC Violations | 0 | 0 | +0 |
| Maximum Curvature (1/m) | 0.0123 | 0.0091 | -26.0% |
| Maximum Lateral Jerk (m/s$^3$) | 8.6400 | 5.4146 | -37.3% |
| RMS Lateral Jerk (m/s$^3$) | 4.0199 | 2.4689 | -38.6% |

**Analysis of quantitative metrics.**

Table 6 provides a direct numerical comparison between the baseline quintic and the proposed TTC-integrated planner. Several key observations can be drawn.

First, the minimum time-to-collision value of the proposed method (2.963s) is slightly lower than that of the baseline (3.287s). This is not due to reduced safety—in both cases the number of TTC violations is zero—but rather because the stable method proactively exploits the earliest feasible lane-change window once the TTC threshold is satisfied. In other words, the algorithm adopts a more decisive trajectory that remains safe but avoids unnecessary delay, leading to a numerically smaller minimum TTC while still maintaining safety-guaranteed operation.

Second, passenger comfort is significantly improved. The maximum curvature is reduced by 26%, while both the peak and RMS lateral jerk are reduced by more than 35%. This indicates that the TTC-penalized optimization not only ensures safety but also systematically suppresses oscillations and abrupt motions that degrade ride quality. The smoother lateral jerk profile is also evident in Fig. 30.

Overall, the results confirm that the proposed framework achieves a balanced trade-off: it selects the earliest safe lane-change opportunity (hence a slightly smaller minimum TTC), while delivering smoother and more comfortable trajectories with substantially reduced jerk and curvature.

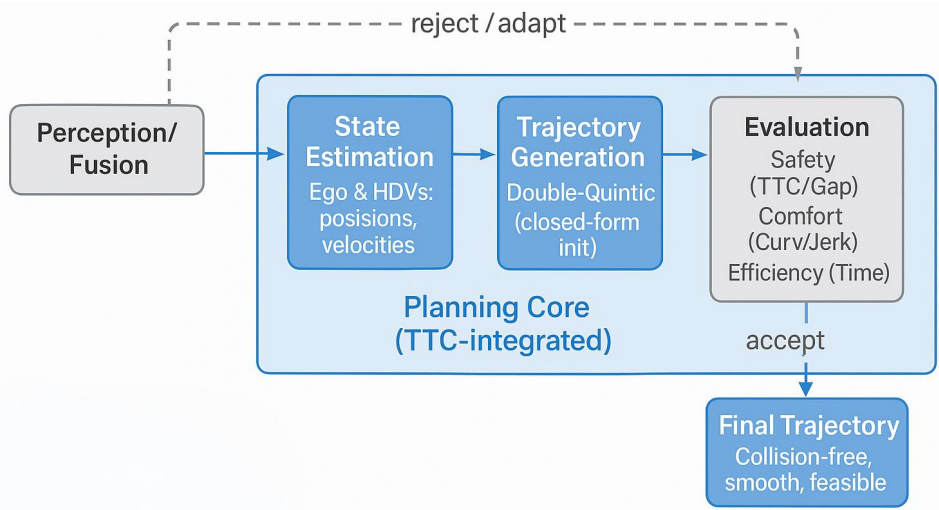

**Fig. 31** Overall workflow of the proposed framework, illustrating the sequence from state estimation, trajectory generation, TTC computation, and evaluation, to the final safe and comfortable trajectory. A feedback loop ensures adaptive adjustment based on evaluation results.

## 7 Conclusion

### 7.1 Scope and Limitations

While the proposed TTC-integrated double-quintic framework demonstrates strong performance across lane-changing and overtaking scenarios, it is important to clearly delineate its scope and limitations. The method is best suited for structured highway

This paper presented an improved double quintic polynomial approach for safe and efficient lane-changing in mixed traffic environments. By integrating a TTC based evaluation mechanism into the trajectory generation process, the proposed framework ensures that the ego vehicle maintains a safe distance from surrounding HDVs throughout the maneuver. The approach combines environment perception, state estimation, trajectory generation, and real-time safety evaluation to adaptively plan safety-guaranteed lane changes while ensuring passenger comfort. Extensive simulations conducted under various traffic scenarios demonstrated the safety, effieicncy, and comforts of the proposed method over conventional trajectory planning approaches, including quintic polynomials, Bezier curves, and B-splines. The results showed that the improved double quintic method not only enhances safety but also achieves smoother and more efficient transitions, outperforming existing solutions in both static and dynamic environments. Future work will focus on addressing uncertainties in the driving environment. Real-world traffic conditions are inherently unpredictable, with variability in human behavior, sensor noise, and environmental factors.Furthermore, future research will focus on extending the proposed framework to more realistic vehicle dynamics by incorporating steering constraints, acceleration / deceleration limits, and sideslip effects. These improvements will enhance the applicability of the method to safety-critical high-speed lane-changing scenarios, ensuring that the planned trajectories remain feasible under realistic vehicle performance. At last, Although the experimental evaluation was conducted on a desktop platform equipped with an RTX 3070 Ti GPU, the proposed method is inherently lightweight. Unlike deep learning-based planners that require extensive parallel computing, our framework relies on polynomial trajectory generation with an embedded TTC penalty. The optimization converges reliably within 20 iterations, which indicates that the computational burden is modest and suitable for real-time execution. Furthermore, the closed-form initialization (Eq. 17) and matrix-based formulation keep the complexity tractable, avoiding the need for large-scale training or GPU-dependent operations. These characteristics make the method well-suited for deployment on resource-constrained embedded devices such as NVIDIA Jetson AGX, where real-time feasibility is critical. While this work primarily validated the algorithm in simulation, future efforts will extend to hardware-in-the-loop experiments on embedded platforms to further confirm its efficiency and practicality.

**Supplementary information.** If your article has accompanying supplementary file/s please state so here. Authors reporting data from electrophoretic gels and blots should supply the full unprocessed scans for key as part of their Supplementary information. This may be requested by the editorial team/s if it is missing.

Please refer to Journal-level guidance for any specific requirements.

**Acknowledgements.** Acknowledgements are not compulsory. Where included they should be brief. Grant or contribution numbers may be acknowledged. Please refer to Journal-level guidance for any specific requirements.

# Declarations

Some journals require declarations to be submitted in a standardised format. Please check the Instructions for Authors of the journal to which you are submitting to see if you need to complete this section. If yes, your manuscript must contain the following sections under the heading 'Declarations':